\documentclass[single-column]{fairmeta_custom}

\usepackage{comment}
\usepackage{amsfonts}

\newif\ifshowcomments

\showcommentstrue

\definecolor{julieta_colour}{RGB}{217,95,2} %

\newcommand{\todo}[1]{\noindent{\color{red}{\bf TODO:} {#1}}}

\ifshowcomments
    \newcommand{\yk}[1]{ \noindent {\color{gray} {#1} -- Yash} }
    \newcommand{\ai}[1]{ \noindent {\color{gray} {\bf ChatGPT:} {#1}} }
    \newcommand{\tb}[1]{ \noindent {\color{blue} {\bf Timur:} {#1}} }    
    \newcommand{\jm}[1]{ \noindent {\color{julieta_colour} {\bf Julieta:} {#1}} }    
 \else
    \newcommand{\yk}[1]{\unskip}
    \newcommand{\todo}[1]{\unskip}
    \newcommand{\ai}[1]{\unskip}
    \newcommand{\tb}[1]{\unskip}        
    \newcommand{\jm}[1]{\unskip}        
\fi

\newcommand{\heading}[1]{\noindent\textbf{#1}}

\def\ie{i.e.,\ }               %

\def\ig{Humans-3B\ }
\def\fullbody{Full-body\xspace }
\def\upperbody{Head-only\xspace }
\def\mobile{iPhone\xspace }
\def\ourmodel{Pippo\xspace }

\def\spatialanchor{Spatial Anchor\xspace}

\def\plucker{Plücker\xspace }

\def\resone{$128 \times 128$\ }
\def\restwo{$256 \times 256$\ }
\def\resthree{$512 \times 512$\ }
\def\resfour{$1024 \times 1024$\ }

\def\posttraining{Post-training\xspace }
\def\post{P$3$}
\def\midtraining{Mid-training\xspace }
\def\mid{M$2$}
\def\pre{P$1$}

\def\reprojectionerror{Reprojection Error\xspace}

\def\noise{\bm{\epsilon}}

\def\dmmodel{\epsilon_\theta}

\def\degree{^{\circ}}

\definecolor{demphcolor}{RGB}{100,100,100}

\newcommand{\STAB}[1]{\begin{tabular}{@{}c@{}}#1\end{tabular}}

\definecolor{cvprblue}{rgb}{0.21,0.49,0.74}

\crefname{section}{Sec.}{Secs.}
\Crefname{section}{Section}{Sections}
\Crefname{table}{Table}{Tables}
\crefname{table}{Tab.}{Tabs.}
\Crefname{figure}{Figure}{Figures}
\crefname{figure}{Fig.}{Figs.}

\title{Pippo: High-Resolution Multi-View Humans from a Single Image}

\author[1,2]{Yash Kant}
\author[1,3]{Ethan Weber}
\author[1]{Jin Kyu Kim}
\author[1]{Rawal Khirodkar}
\author[1]{Su Zhaoen}
\author[1]{Julieta Martinez}
\author[2\dagger]{\\ Igor Gilitschenski}
\author[1\dagger]{Shunsuke Saito}
\author[1\dagger]{Timur Bagautdinov}

\affiliation[1]{Meta Reality Labs}
\affiliation[2]{University of Toronto}
\affiliation[3]{UC Berkeley}

\contribution[\dagger]{Joint Advising}

\abstract{\begin{abstract}

We present \ourmodel, a generative model capable of producing 1K resolution dense turnaround videos of a person from a single casually clicked photo.  
\ourmodel is a multi-view diffusion transformer and does not require any additional inputs — e.g., a fitted parametric model or camera parameters of the input image.
We pre-train \ourmodel on 3B human images without captions, and conduct multi-view mid-training and post-training on studio captured humans. 
During mid-training, to quickly absorb the studio dataset, we denoise several (upto 48) views at low-resolution, and encode target cameras coarsely using a shallow MLP. 
During post-training, we denoise fewer views at high-resolution and use pixel-aligned controls (e.g., Spatial anchor and Plucker rays) to enable 3D consistent generations. 
At inference, we propose an attention biasing technique that allows \ourmodel to simultaneously generate greater than 5$\times$ as many views as seen during training.
Finally, we also introduce an improved metric to evaluate 3D consistency of multi-view generations, and show that \ourmodel outperforms existing works on multi-view human generation from a single image. Webpage: \url{http://yashkant.github.io/pippo}

\end{abstract}
}

\begin{document}

\maketitle

\section{Introduction}

Creating photorealistic human representations with the ability to control viewpoints has numerous applications in entertainment, healthcare, fashion and
social media. 
Building such representations, first and foremost, requires high-quality multi-view studio data~\cite{isik2023humanrf, bagautdinov2021driving}, which is costly to acquire.
This significantly limits the scalability in terms of the number identities for high-quality studio data~\cite{isik2023humanrf, martinez2024codec, xiong2024mvhumannet}.
In contrast, large-scale, unstructured, and diverse human images and videos are available online. 
However, the raw data of such in-the-wild images does not offer ground-truth 3D or multi-view representations of humans. 

\begin{figure}[t!]
    \centering
    \includegraphics[width=\textwidth]{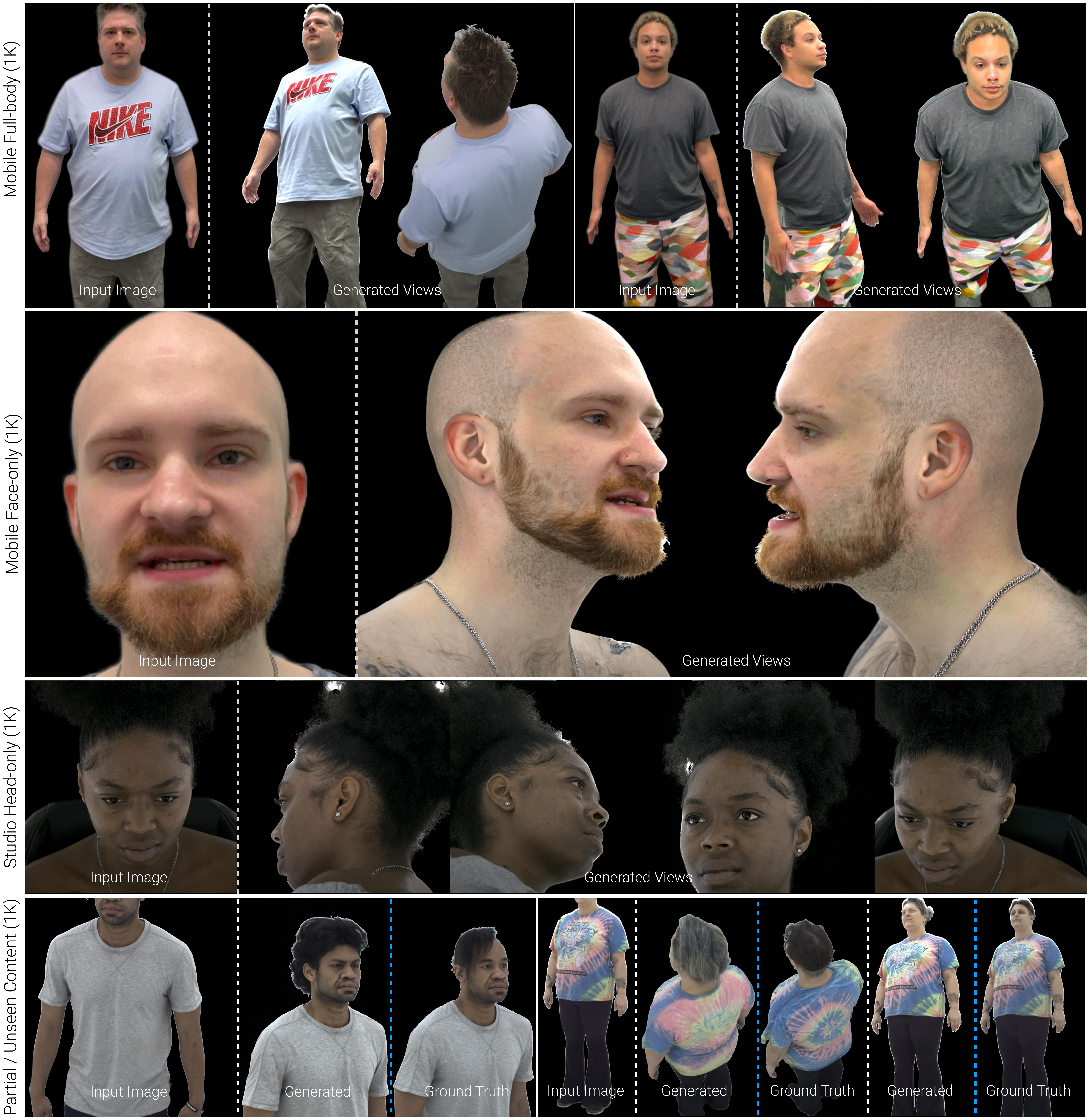}
    \captionof{figure}{\textbf{\ourmodel generates high-resolution, multi-view, studio-quality images from a single photo.} In each sample, the left-most image is the input, followed by novel generated views of unseen subjects. First and second rows show generations from Full-body and Face-only photos captured in-the-wild using a mobile phone. Third row shows generation from a Head-only studio image. Last row illustrates \ourmodel's capability to faithfully blend observed and generated content, alongside the corresponding ground truth.} 
    \label{fig:teaser}
\end{figure}

In this work, we present a novel approach leveraging the best of two worlds: 
generalizability from in-the-wild unstructured images, and fidelity and view-controllability from studio capture data. 
Specifically, our model \ourmodel is a diffusion transformer, which can generate several 1K resolution multi-view consistent images jointly during inference.
\ourmodel takes as input a single image of an individual, camera poses of the target viewpoints to be generated. Since the scale and placement of the subject is ambiguous from a single image, \ourmodel uses a Spatial Anchor which roughly specifies the location and orientation of the subject in 3D space.
Our model does not rely on additional conditioning such as body priors or
camera parameters of the input images, to scale our training pipeline to in-the-wild data and support unconstrained inputs at test time.

We employ a multi-stage training recipe to train \ourmodel. First, we pre-train the model for latent-to-image generation task, similar to ~\cite{DALLE-2}, on a large human-centric dataset of in-the-wild images.
Next, in \midtraining stage, we jointly generate multiple consistent images of the subject, conditioned on target viewpoints and a single input image using high-quality studio dataset. 
Finally, in \posttraining stage, the model is provided with a minimal 
placement signal - \textit{spatial anchor} - encoding rough head orientation - 
which further improves 3D consistency.
Our architecture is also carefully tailored to the conditional multi-view generation - we propose several 
simple but effective modifications to the basic DiT architecture, including self-attention-based conditioning,
lightweight spatial controls and camera conditioning with \plucker coordinates.

During inference, our goal is to produce smooth turnaround videos, which requires generating up to five times more views than were present during training. However, we observe that simply increasing the number of views leads to a drop in quality of the generated content. Our investigation reveals that this drop is due to heightened entropy in the attention heads as the number of views increases. To address this problem, inspired by previous research in super-resolution~\cite{jin2023training}, we introduce an attention biasing approach to control and reduce entropy growth within multi-view models.

For evaluating such a multi-view diffusion model, 3D consistency is a critical metric for understanding the geometric correctness of the generation. 
For generation task in an ill-posed setup (e.g., a single image input), there may be multiple possibilities that are equally plausible. 
Existing multi-view generation methods typically report reconstruction metrics (\ie PSNR, SSIM and LPIPS)~\cite{gao2024cat3d} 
or FID~\cite{zero1to3}, which either (1) penalize any new content that is actually 3D consistent or 
(2) are unable to measure the 3D consistency of the generation. 
To address this problem, we design a metric measuring 3D consistency: we compute 2D keypoint matches
with an off-the-shelf method \cite{sarlin2020superglue}, triangulate them, and then reproject back into the other views to
measure the reprojection error in pixels.
Some generation methods measure reprojection errors~\cite{fridman2024scenescape} from SfM~\cite{SFM} or epipolar distance~\cite{muller2024multidiff,TrainNVSDM3}, but our metric uses camera pose as input and is more precise than measuring distance to a epipolar line. We show our metric helps to quantify our results, finding that our method is favorable compared with existing approaches and other baseline methods.

\noindent To summarize, we make the following contributions:
\begin{itemize}
    \item a generative model capable of generating high-resolution and multi-view consistent humans from a single image and its effective training strategy.
    \item a diffusion transformer architecture designed to enhance multi-view generation and viewpoint control.

    \item an attention biasing technique to enable generating >5\texttt{x} more views at inference compared to training.

    \item a novel 3D consistency metric to accurately measure the level of 3D consistency in generative tasks. 
\end{itemize}

\section{Related Work}
\label{sec:related_works}
\begin{figure*}[t!]
    \centering    
    \includegraphics[width=\textwidth]{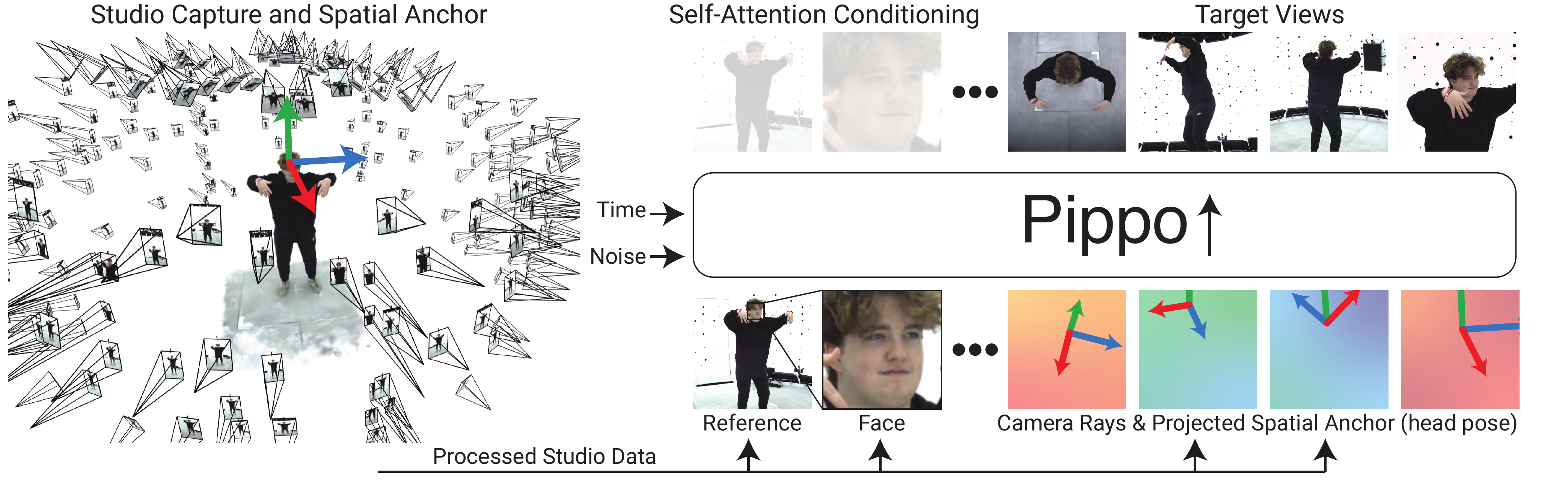}
    \captionof{figure}{\textbf{Pipeline overview}. This is an illustration of how we train our model. (Left) we use data from a studio capture and train our multi-view diffusion model (right). We condition on a full reference photo and a cropped face, as well as the target view cameras and 2D projected spatial anchor indicating head position and orientation. Our diffusion model also takes in noisy target views and a timestep in order to predict the denoised views (top). In practice, we apply a segmentation mask around the person.} 
    \label{fig:pipeline}
\end{figure*}

We review multi-view human datasets and generative models for human synthesis,  categorizing methods by their data requirements and prior constraints—such as parametric human models or explicit 3D structures. In this work, we minimize reliance on complex priors by training the model on large amounts of human-centric data and impose minimal camera and spatial 
controls.

\vspace{1mm}
\heading{Multi-view Human Datasets.}
While many large 2D human datasets exist~\cite{lin2019coco, LAION, ImageNet}, 3d captures of people (\ie captures using dozens of views at a time) are still relatively rare, since they are expensive to collect and are thus mostly obtained in specialized research labs.
Nevertheless, this kind of data has recently become more common, going from hundreds to thousands of publicly-available captures.
For face-only captures, a few datasets~\cite{Yu_2020_CVPR,yang2020facescape,yenamandra2020i3dmm,2023renderme360,martinez2024codec} provide up to 600 subjects captured from over 60 views each,
while for full-body captures, 4D-Dress~\cite{wang20244ddress} provides 32 subjects with 2 outfits each, and detailed annotations for their garments, while DNA-rendering~\cite{2023dnarendering} provides 500 subjects from 60 views.
MVHumanNet~\cite{xiong2024mvhumannet} stands apart by providing 4,500 people captured from 48 views in and over 9,000 sets of clothing, outpacing other datasets by an order of magnitude in terms of people diversity.
We use internal multi-view data of $~\sim$1000 identities with dense ($\sim$160) 
camera setup, but we expect our model to produce reasonable results
with publicly-available captures.

\vspace{1mm}
\heading{Generating Novel Poses and Expressions.}
Although different from our task, this related work builds 3D animatable models of people and faces. These methods use neural radiance fields~\cite{HumanNerf, su2023npc, su2022danbo}, or 3d Gaussians~\cite{zielonka2023drivable, qian20243dgs, wen2024gomavatar,kirschstein2023nersemble}, using canonical ``T''-poses and a corresponding forward and backward mapping to model local appearance changes.
Many of these methods, however, rely on accurate human shape and pose estimation, and only achieve high photorealism for personalized models from studio captures. 
Relatedly, a family of methods focus on generating animatable faces from a single image by disentangling viewpoints and expressions from large collections of 2d portraits~\cite{megaportraits, drobyshev2024emoportraits, zhang2023metaportrait, xu2024vasa, Nerfies, StyleGAN}, allowing photorealistic realtime reenactment, while being limited to faces and small viewpoint variations.
Unlike these methods, we complete missing parts of a person given either one or a few partial views.
We do not repose the person or animate the faces but can instead recover entirely missing views, which is orthogonal to these methods.
Moreover, our approach addresses both full-body and facial reconstruction, rather than specializing in either domain.

\heading{Generations with Explicit 3D Structures.}
Similar to early methods that extracted 3d representations from 2d models via score distillation~\cite{dreamfusion,TextMesh,fantasia3d,Nerdi,ProlificDreamer,magic3d,dreambooth3d}, 3d representations of human appearance have been extracted from GANs by co-training a neural renderer from learned TriPlane~\cite{EG3D} or HexPlane~\cite{An_2023_CVPR, li2025spherehead} latent spaces.
Others train generative models that operate directly in 3D space with diffusion or GAN objectives~\cite{Rodin3DDM,muller2023diffrf,TrainOn3D1,TrainOn3D2,TrainOn3D4,TrainOn3D5} or regression models to directly predict 3D representations~\cite{PiFU,saito2020pifuhd,sengupta2024diffhuman}.
While these methods produce 3d consistent faces by design, their quality is limited by the relatively small amounts of 3d training data
and/or the limitations of their respective 3d representations.
In contrast, we focus on generating 3d-consistent 2d images and thus avoid the downsides of explicit 3d modelling.

\heading{Generations with 2D Models.}
Another option is to model viewpoint changes in 2d, without an underlying 3d representation~\cite{zero1to3}, sometimes modelling either photometric~\cite{iNVS} or epipolar~\cite{huang2024epidiff} constraints explicitly.
DiffPortrait3D~\cite{gu2024diffportrait3d} is a portrait-generation method that falls roughly within this category, by fine-tuning a 2d diffusion model for 3d-aware face generation using ControlNet-style~\cite{zhang2023adding} conditioning for viewpoint control, as well as cross-view attention~\cite{guo2023animatediff} and initialization from a 3d-consistent GAN~\cite{EG3D}.
Other methods are more general and not specific to humans, focusing on single-view to 3D generation~\cite{zero1to3,NeuralLift-360,sargent2023zeronvs,seo2024genwarp,Make-It-3D,shen2023anything3d,gao2024cat3d}. In addition to these single-view methods, video models have been fine-tuned for camera control~\cite{he2024cameractrl,wang2024motionctrl}.

\section{Method}\label{sec:method}

We train our models following a three-stage strategy:

\begin{itemize}
    \item \textbf{Image-only Pre-training (P1).} We pretrain on a large-scale human-centric dataset with image conditioning.
    \item \textbf{Multiview Mid-training (M2).} We train models at a low-resolution of \resone to denoise $48$ target views with coarse camera control (no pixel-aligned spatial control).
    \item \textbf{Multiview Post-training (P3).} We train at a high-resolution of \resfour to denoise $1-3$ target views with spatial control injected via ControlMLP layers. 
\end{itemize}

\noindent We denote the training stage and resolution of any given model as \{stage\}@\{resolution\}. For instance, M2@128 represents a mid-trained model at 128 resolution.

\subsection{Base Model} \label{ssec:base_model}

\heading{Architecture.} We adopt a DiT-like~\cite{peebles2023scalable} architecture with scale, shift, and gate modulation for timestep 
conditioning inspired by Stable Diffusion 3~\cite{esser2024scaling} and Flux~\cite{flux}.
We simplify the architecture by employing MLP and attention in parallel~\cite{zhai2022scaling}, and removing second LayerNorm after 
attention layers.
We use 
VAE for training in latent space with $8$x spatial compression, and patchify latent images via a linear layer and patch size of $2$. We use fixed sinusoidal positional encoding during training. We provide more details in~\cref{app_sec:dit}, and the exact design of our DiT block in~\cref{fig:dit_block}. 

\heading{Image-only Pre-training.} During pre-training, the model learns to 
denoise an image conditioned on its corresponding image embedding from 
DINOv2~\cite{darcet2023vitneedreg,oquab2023dinov2},
this is similar in principle to the image decoder of DALL-E 2~\cite{DALLE-2}.
We project both embeddings to the model dimension using with a linear layer 
to create a joint conditioning. Importantly, our pretraining setup does not require any 
annotations or captions for the images, and is well-aligned with our downstream 
objective of generating consistent multiview images given a single reference image as 
input.

Formally, given an image $\mathbf{y} \in \mathbb R^{H \times W \times C}$, and the joint conditioning as $\mathbf{e}^{\text{img}} \in \mathbb R^{N \times D}$. We pre-train our diffusion model $\epsilon_\theta$ with the following objective: 
\begin{equation}\label{eq:dm_loss}
    \mathcal{L}_{\mathrm{DM}} = ||\noise^t - \dmmodel(\mathbf{y}^{t}, \mathbf{e}^{\text{img}},  t)||^2
\end{equation}

\noindent Where $t \in \mathbb [0,T]$ is diffusion timestep, $\noise^t \sim \mathcal{N}(\bm{0}, \bm{I})$ is the noise added at the given timestep.  We use the DDPM~\citep{DDPM,FirstDiffusionModel} formulation to define discrete timesteps and set $T=1000$. We first pretrain our model at \restwo and then at \resthree on a large corpus of human-centric images. Exact training details can be found in~\cref{app_sec:diffusion}.

\subsection{Multiview Model} \label{ssec:multiview_model}

Our goal is to generate many high resolution and unseen novel viewpoints of a human subject (akin to a studio capture) given a single input image.

\vspace{1mm}
\heading{Input Reference.} We denote the input image as $\mathbf{x}^{\text{ref}}$ and corresponding face crop as $\mathbf{x}^{\text{face}}$. The face crop is obtained using FaceNet~\cite{schroff2015facenet} and resized to the same size as input image, such that: $\mathbf{x}^{\text{face}}, \mathbf{x}^{\text{ref}} \in \mathbb R^{H \times W \times C}$. 

\vspace{1mm}
\heading{Target Cameras.} We denote the target viewpoints to be synthesized using distinct cameras (intrinsics and extrinsics), represented as $\mathbf{c}_{1:N}$. Each camera
is used to generate its \plucker coordinates $\mathbf{P}_i \in \mathbb R^{H \times W \times 6}$.

\vspace{1mm}
\heading{Target \spatialanchor.} In addition to target cameras, we provide an oriented 3D point denoted as $\mathbf{a}_i=[\mathbf{R}_{i}|\mathbf{t}_{i}]$, which roughly defines the center of the subject's head, as well as their gaze direction. We show an example of our \spatialanchor in~\cref{fig:pipeline} in the studio on the left. This anchor is color-coded and projected into a 2D image, which is used as conditioning for our target view generations.  During inference, the \spatialanchor can be placed at any given point that lies within the field-of-view of the target cameras.

\vspace{1mm}
\heading{Multiview Diffusion Model.} Given the above inputs, we train a multiview diffusion model $\epsilon_\theta$ to \emph{jointly} denoise all the target views $\mathbf{y}_{1:N} \in \mathbb R^{N \times H \times W \times C}$ with the objective: 
\begin{equation}\label{eq:dm_loss2}
    \mathcal{L}_{\mathrm{DM}} = ||\noise^t - \dmmodel(\mathbf{y}^{t}_{1:N}, \mathbf{c}_{1:N}, \mathbf{x}^{\text{ref}}, \mathbf{x}^{\text{face}},  t)||^2
\end{equation}

\noindent where 
$\mathbf{y}^{t}_{1:N}$ are noisy target images. We condition the base model on the provided reference image and its face crop by concatenating their patchified latent tokens with the noisy input latent tokens to the model. This is shown in ~\cref{fig:pipeline}

\begin{figure}[t]
    \centering    
    \includegraphics[height=7cm, trim=13cm 0 13cm 0, clip]{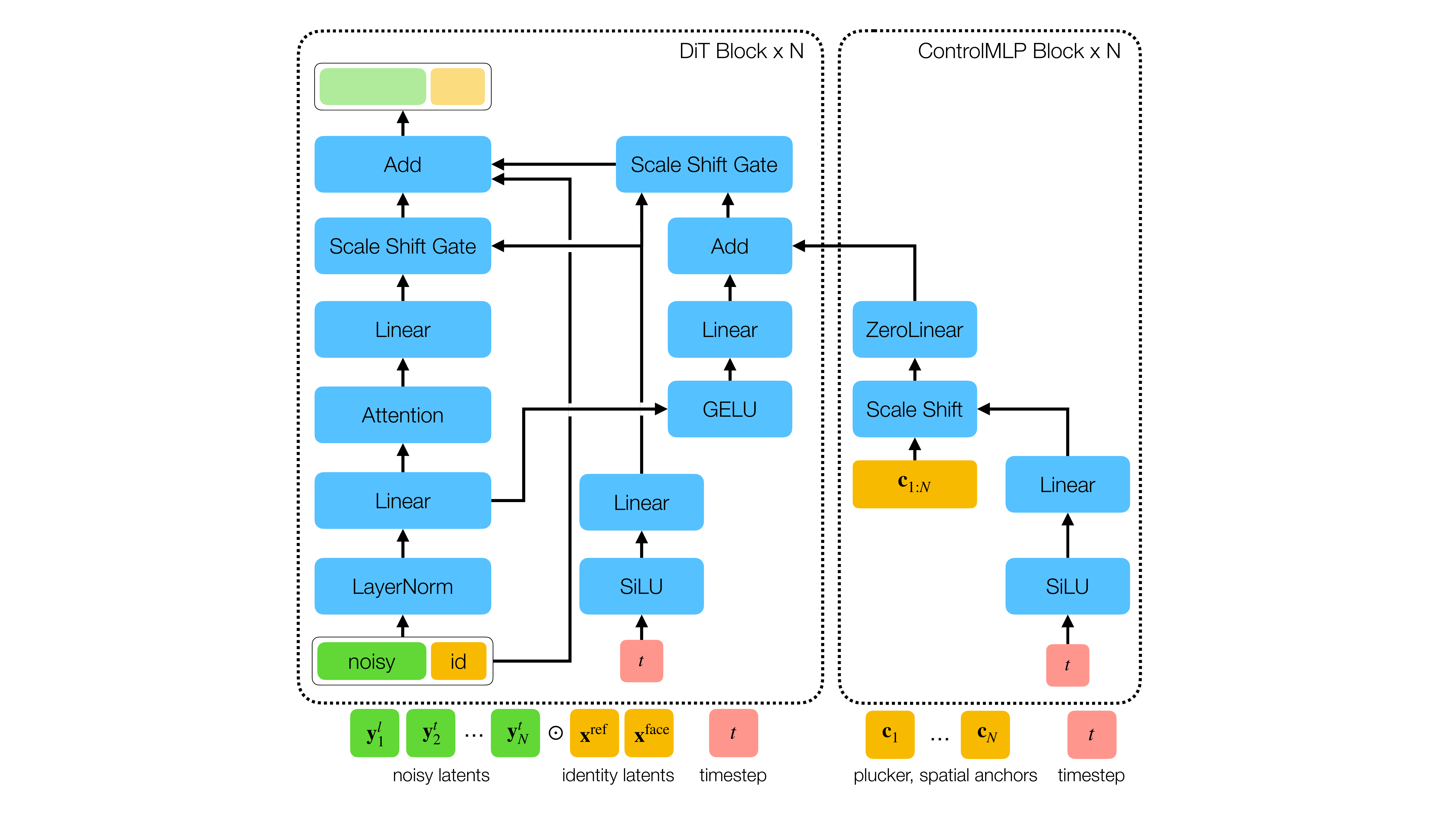}
    \vspace{-0.5cm}
    \caption{\textbf{DiT and ControlMLP Block}. Our DiT block (left) loosely follows~\cite{stable_diffusion}, 
    with a AdaIn-based timestep modulation. 
    We apply attention and MLP blocks in parallel~\cite{pmlr-v202-dehghani23a}, 
    and jointly apply self-attention to the noisy generated and identity conditioning tokens. 
    ControlMLP block (right) is used to provide lightweight spatially-aligned conditioning - 
    \plucker and \spatialanchor. }
    \label{fig:dit_block}
    \vspace{-0.2in}
\end{figure}

\vspace{1mm}
\heading{\midtraining.} In mid-training stage, we want to train a strong multiview model that can denoise several images together, and absorb the dataset quickly at a lower resolution. During this phase, we do not use any pixel-aligned spatial control such as \plucker or \spatialanchor. We use an MLP to encode the flattened $16$-dimensional target camera intrinsics and extrinsics into a single token 
. We fuse this camera token into each noisy latent token (for the corresponding view) as positional encoding, which makes our multiview model 3D-aware of the target viewpoints. 
We mid-train our model at \resone resolution to jointly denoise 24 views.

\vspace{1mm}
\heading{\posttraining.} In the post-training stage, our objective is to create a high-resolution model which is 3D-consistent starting with a low-resolution and a 3D-aware (but not consistent) model. For this, we design a lightweight ControlNet~\cite{zhang2023adding}-inspired module, which takes as input the pixel-aligned \plucker and \spatialanchor controls, and the denoising timestep to create a separate modulation signal for the multiview model. We name this module ControlMLP, as it uses a single MLP to generate scale-and-shift modulated control for each multiview-DiT block as shown in ~\cref{fig:dit_block}. Each layer of ControlMLP is zero-initialized at the start. We find that the Post-training phase is crucial in reducing flicker and 3D inconsistencies in generations. We post-train models at \resthree and \resfour resolutions to jointly denoise 10 and 2 views respectively. Increasing the number of views further lead to GPU out-of-memory issues.

\vspace{1mm}
\heading{Encoding \plucker and \spatialanchor.} We notice that the relative differences between neighboring pixels in \plucker coordinates are tiny. To amplify these differences better, we use a SIREN~\cite{SIREN} layer to first process the $6$D grid into a $32$D feature grid. Then, we downsample it by $8$x to match the size of latent tokens and feed it as input to the ControlMLP. In addition, use the \spatialanchor to fix the position and orientation of the subject's head in 3D. We only use the \spatialanchor for generations, and not for the input reference view. We encode the \spatialanchor image into the latent space of our model via VAE, and concatenate it with \plucker input and pass it through an MLP to create the modulation signal at each layer.

\subsection{Understanding and Improving Spatial Control}\label{ssec:spatial_control}

\noindent This section presents our design choices and examines alternative approaches for injecting pixel-aligned spatial controls during \posttraining stage. We demonstrate the effectiveness of spatial control through a focused overfitting experiment with quantitative evaluations in ~\cref{tab:3deval}.

\vspace{1mm}
\heading{Scene Overfitting Task.} We use $160$ frames from a fixed 3D scene of a given subject and timestamp, split into $100$ training and $60$ validation views. We overfit our mid-trained model to the training views while testing various spatial control methods, training only the control modules while keeping other weights frozen. After overfitting for $10$K iterations, we evaluate the model on validation views for novel view synthesis.  Strong generalization to validation viewpoints indicates effective spatial control and appropriate camera viewpoint sensitivity. Through this task, we evaluate different spatial control injection methods in~\cref{tab:3deval}, starting with simple to advanced modulation designs.

\begin{table}[h!]
    \centering
    \setlength{\tabcolsep}{5pt}
    \small
    \begin{tabular}{llcc}
        \toprule
         {\#} & \textbf{Method (\mid @ 128)} & PSNR$_{\text{val}}$ $\uparrow$ & PSNR$_{\text{train}}$ $\uparrow$\\

    \midrule
        {1.} & Mid-trained (No Overfitting) & 19.23 & 19.70 \\
    \midrule
    
    {2.} & + Camera (w/ MLP)~\cite{zero1to3, MVDream} & 17.95\textsubscript{{\color{red}{-1.28}}} & 19.92\textsubscript{{\color{green}{+0.22}}} \\
    
    {3.} & + \plucker (w/ MLP)~\cite{RayConditioningGAN,LFNs,kant2024spad,he2024cameractrl}   & 18.89\textsubscript{{\color{red}{-0.34}}} & 20.74\textsubscript{{\color{green}{+1.04}}} \\
    
    {4.} & + ControlMLP  & 19.45\textsubscript{{\color{green}{+0.22}}} & 29.36\textsubscript{{\color{green}{+9.66}}} \\
    
    {5.} & + SIREN  & 20.13\textsubscript{{\color{green}{+0.90}}} & 30.19\textsubscript{{\color{green}{+10.49}}} \\
    
    {6.} & + \spatialanchor (Ours) \ & 22.60\textsubscript{{\color{green}{+3.37}}} & 30.49\textsubscript{{\color{green}{+10.79}}} \\
    \bottomrule

    \end{tabular}
    \caption{
        \textbf{Evaluating and Designing Spatial Controls (\cref{ssec:spatial_control}).} We overfit our multi-view model for $10$K iterations on 100 views of a single scene from our \upperbody dataset, and evaluate on 60 novel views of this scene by only varying the spatial controls (camera and \spatialanchor). We use this setup to test the modulation strength of different spatial controls. Subscript values \textcolor{green}{green}/\textcolor{red}{red} show deviations from Row 1. }
    \vspace{-10pt}
    \label{tab:3deval}
\end{table}

\begin{itemize}
    \item \textbf{No overfitting (Row 1).} The Mid-trained model without scene-specific overfitting achieves comparable PSNR of 19.2 and 19.7 on train and validation views respectively. We treat this setup as baseline to improve over. 

    \item \textbf{Encoding Camera with MLP (Row 2).} We encode camera using an MLP similar to prior works~\cite{zero1to3, MVDream} and our \midtraining stage (\cref{ssec:multiview_model}). After overfitting, the model achieves slightly better PSNR on training views as expected, however the validation PSNR drops by 1.28 points to 17.95. This suggests that an MLP does not provide enough modulation for camera control. 
    
    \item \textbf{\plucker as Positional Encoding (Row 3).} In this setup, we use downsampled and patchified \plucker coordinates processed through MLP to create positional encoding, which is added to the noisy latent tokens. This setup is inspired from  prior works~\cite{RayConditioningGAN,LFNs,kant2024spad,he2024cameractrl,bahmani2024vd3d}, and it further improves the validation PSNR compared to MLP at 18.89, but lags behind the non-overfitted baseline.

    \item \textbf{\plucker with ControlMLP and SIREN (Row 4, 5).} Here, we use our ControlMLP module to inject spatial control at each multiview-DiT block output. Moreover, encoding \plucker coordinate with a SIREN~\cite{SIREN} amplifies the relative differences between neighboring pixels (\cref{ssec:multiview_model}). This setup achieves PSNR of 20.13 with an improvement of 0.9 over baseline.

    \item \textbf{Adding \spatialanchor (Row 6).} Finally using the \spatialanchor gives validation PSNR of 22.6 (gain of 3.3 points over baseline) and enables strong spatial control. Thus, we adopt this configuration for \posttraining stage.
\end{itemize}

\subsection{Handling Varying Number of Views at Inference}\label{ssec:scaling_views}

As discussed in~\cref{ssec:multiview_model}, during training we jointly denoise a fixed number of views. Specifically, 24 views for mid-training at \resone, and 2 or 12 views for post-training at resolutions \resthree and \resfour respectively. This choice is largely motivated to avoid GPU out-of-memory errors during training. However, during inference, we wish to scale the number of views much further to generate smooth turnaround videos. This is feasible because we can run inference at half precision (using bfloat16) and do not need the backprop computation graph to be stored. 

\begin{figure}[h!]
    \centering    
    \hspace{-0.5cm}
    \includegraphics[width=\linewidth]{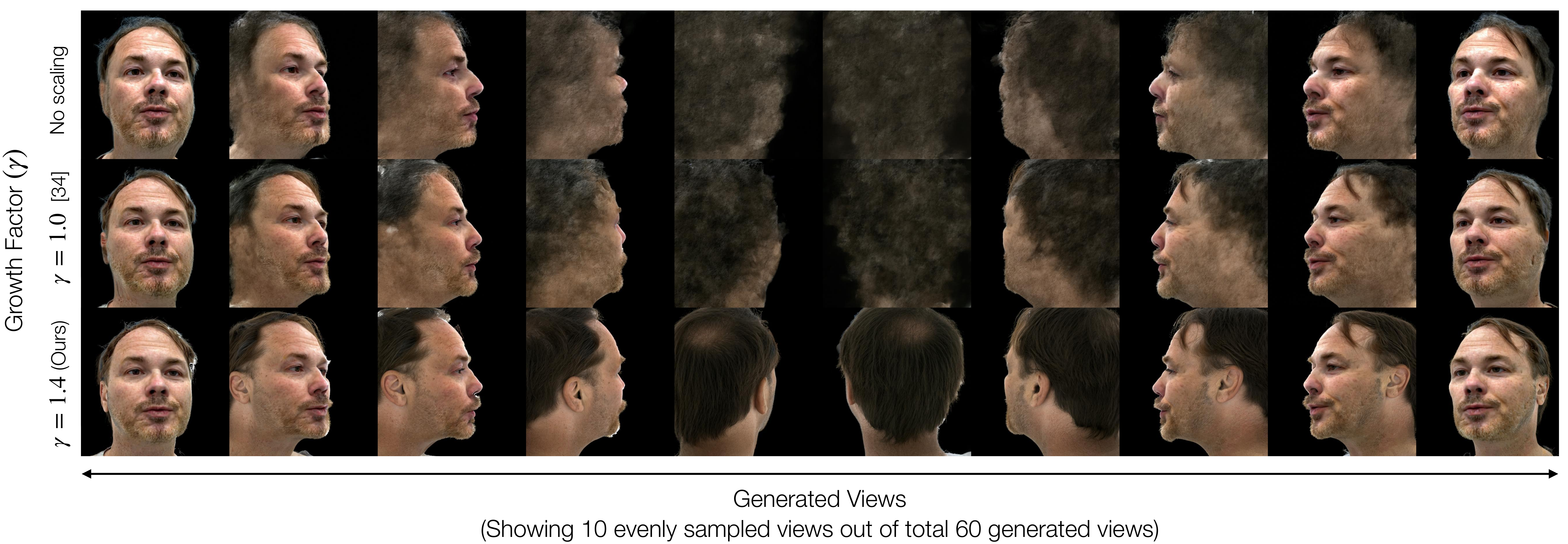}
    \vspace{-0.2cm}
    \caption{{\textbf{Generations under varying strengths of growth factor $\gamma$ (\cref{ssec:scaling_views})}. On each row we show the generated views across vanilla attention~\cite{vaswani2017attention} (No scaling), prior work~\cite{jin2023training} and our formulation Eq.~\eqref{attn_bias}. It can be seen that growth factor ($\gamma$) greater than 1.0 is crucial to mitigate the entropy buildup. We show only 10 views per row subsampled evenly from 60 views generated at \resthree resolution. The model was trained to jointly denoised only 12 views ($N_i = 5*N_t$).}}
    \label{fig:qual_entropy}
\end{figure}

We find that simply scaling the number of views (or tokens) during inference beyond 2\texttt{x} of the number of views during training leads to blurry and degraded generations. We find these degradations to be most significant in regions unspecified in the input, for example, the back of the head or ears as shown in~\cref{fig:qual_entropy}. We investigate this issue next, and introduce Attention Biasing to remedy it. 

\begin{figure}[h!]
    \centering    
    \includegraphics[height=8cm]{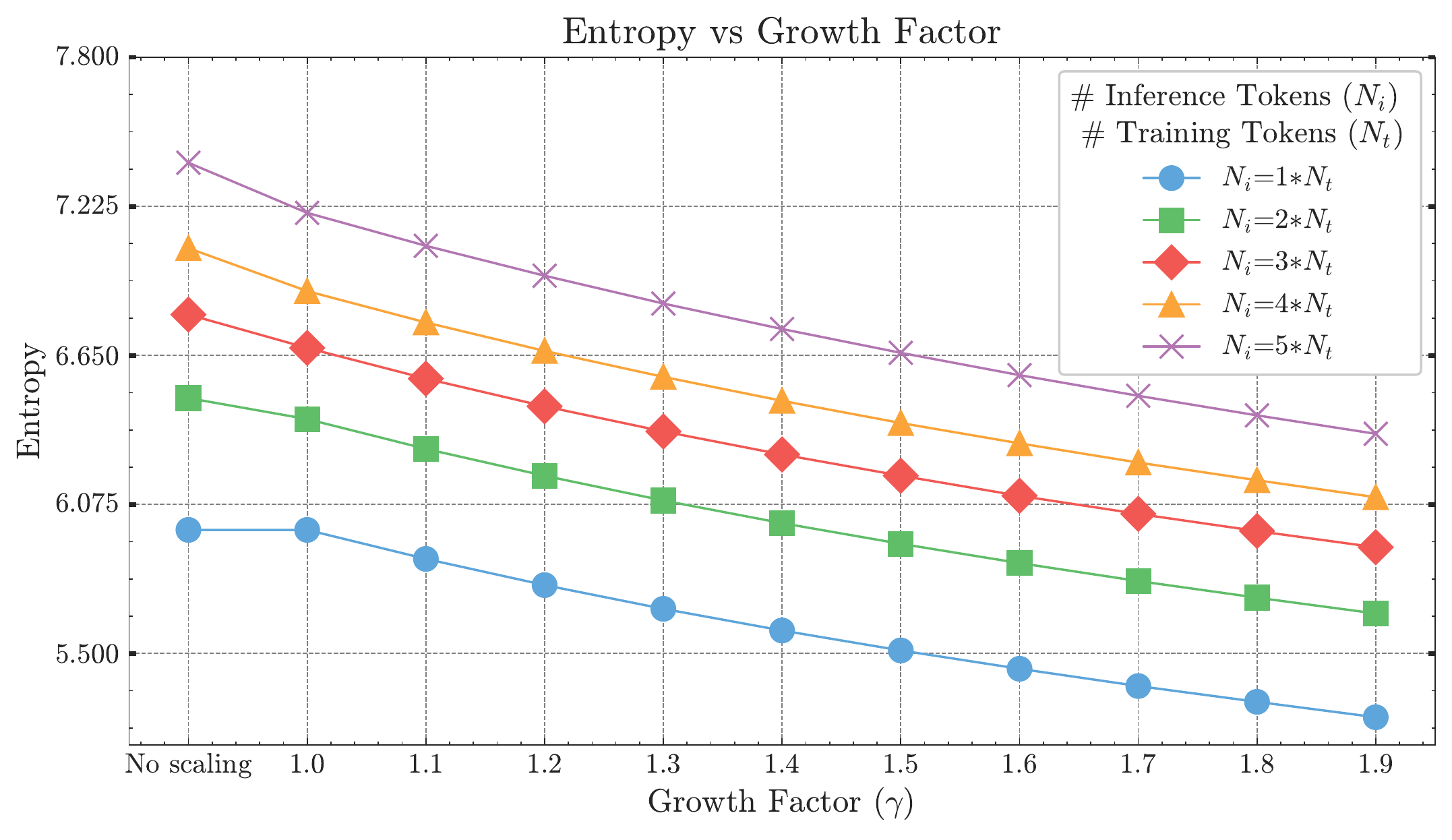}
    \vspace{-0.2cm}
    \caption{{\textbf{Entropy vs Growth Factor ($\gamma$) for varying number of views (tokens) (\cref{ssec:scaling_views})}. We present the entropy results (Y-axis) from our Attention Biasing technique inspired from~\cite{jin2023training} for varying number of tokens (individual line plots), and across different scaling growth factor $\gamma$ introduced in Eq.~\eqref{attn_bias} (X-axis). On X-axis, "No scaling" refers to the default attention formulation~\cite{vaswani2017attention} and $\gamma=1.0$ refers previous work~\cite{jin2023training} formulation. Empirically, we find that a slightly higher value of $\gamma=1.4$  leads to best visuals.}}
    \label{fig:entropy}
\end{figure}

\heading{Attention Biasing.} Let $\textbf{X} \in \mathbb{R}^{N \times d}$ denote the sequence of tokens denoised jointly, where $N, d$ represent number of tokens and the token dimension, respectively. In \ourmodel the total number of tokens in the sequence is proportional to number of views since the VAE and Patchify jointly compress the image by 16\texttt{x} in height and width. Within each DiT block, we compute the $\textrm{Attention}(\textbf{Q}, \textbf{K}, \textbf{V}) = \textbf{A}\textbf{V}$, where  $\textbf{Q},\textbf{K},\textbf{V}$ are query, key, value matrices and $\textbf{A}$ are attention scores computed via taking a row-wise softmax as follows:
\begin{equation}
    \textbf{A}_{i, j} = \frac{e ^{\lambda \textbf{Q}_{i}\textbf{K}_{j}^{\top}}}{\sum_{j' = 1}^{N} e ^{\lambda \textbf{Q}_{i}\textbf{K}_{j'}^{\top}}}, \label{attention map}
\end{equation}
Where the scaling factor $\lambda$ was proposed to be set to $1 / \sqrt{d}$ by the original work~\cite{vaswani2017attention} to stabilize the softmax operation as $d$ increases and avoiding biased (sharp) softmax distributions. We can quantify the sharpness of the attention by computing its entropy. Following previous works in NLP~\cite{ghader2017does,attanasio2022,zhang2024attention}, and in vision~\cite{jin2023training} we define entropy of attention as:
\begin{equation}
    \textrm{Ent}(\textbf{A}_{i}) = -\sum_{j=1}^{N} \textbf{A}_{i, j} \log (\textbf{A}_{i, j}), \label{entropy}
\end{equation}
By substituting the equation ~\eqref{attention map} in equation~\eqref{entropy}, authors of prior work~\cite{jin2023training} meticulously derive that entropy of attention grows logarithmically with number of tokens as follows: \(\textrm{Ent}(\textbf{A}_{i}) \propto \log N\). Furthermore, the authors show that this growth in entropy can be offset during inference by growing the scaling factor $\lambda$ as follows: 
\begin{equation}
    \begin{aligned}
    \lambda = \sqrt{\frac{1}{d}} \cdot \sqrt{\frac{\log N_i}{\log N_t}}
    \end{aligned}
\end{equation}
Where $N_t, N_i$ denote the number of tokens during training and inference respectively. For more details, please refer to Sections 3.1 and 3.2 of the original work~\cite{jin2023training}. Empirically, we find that having a slightly faster growing $\lambda$ alleviates the degradation better. Hence, we propose a hyperparameter $\gamma$ (growth factor) that is tuned between the range $[1.0,2.0]$ to control the growth of $\lambda$ as follows: 
\begin{equation}
    \begin{aligned}
    \lambda_\texttt{ours} = \sqrt{\frac{1}{d}} \cdot \sqrt{\frac{\gamma \cdot \log N_i}{\log N_t}}
    \end{aligned}
    \label{attn_bias}
\end{equation}
We follow the above Equation~\eqref{attn_bias} with our growth factor hyperparameter $\gamma$. In~\cref{fig:entropy}, we plot the entropy (Y-axis) during an inference pass of \ourmodel across varying number of views (tokens) being denoised and demonstrate the corresponding growth in entropy. Furthermore, we also show the attenuation in the entropy under increasing growth factors $\gamma$ (X-axis). 

We put generated visuals before and after using the suggested attention biasing in~\cref{fig:qual_entropy}. We put visuals sweeping over more values of the growth factor in Appendix~\cref{fig:qual_entropy_full}. Similar techniques have been explored in LLMs for handling and generating text at longer contexts~\cite{peng2023yarn,veličković2024softmax}, where the scaling factor $\lambda$ mentioned above is analogous to the inverse of the temperature scale. 

Additionally, we also found that using a bump function instead of constant Classifier-free Guidance during generation leads to fewer artifacts. We discuss this trick further in~\cref{app_sec:diffusion}.

\subsection{Enhanced 3D Consistency Metric}
\label{ssec:consistencymetric}

Traditionally, the 3D consistency of multiview generation models is evaluated using 2D image metrics such as PSNR, LPIPS, and SSIM against a fixed set of ground truth images. However, this approach unfairly penalizes models that generate plausible and 3D-consistent novel content deviating from the fixed ground truth images.
Some works try to address this by measuring SfM~\cite{SFM,fridman2024scenescape} or epipolar error~\cite{muller2024multidiff,TrainNVSDM3}, but these methods solve for pose or are not robust since they measure against the entire epipolar line. To address these limiations, we use our GT camera poses as input and compute \reprojectionerror (RE) given our known camera poses and predicted correspondences.

Our computation of RE involves the following steps:
\begin{enumerate}
    \item \textbf{Landmarks and Correspondence estimation.} We use SuperPoint~\cite{detone2018superpoint} to detect landmarks in the generated images and employ SuperGlue~\cite{sarlin2020superglue} to establish pairwise correspondences between landmarks across images.
    \item \textbf{Triangulation.} Given the correspondences and camera parameters, we apply Triangulation based on Direct Linear Transformation (DLT)~\cite{Hartley2004} to obtain the corresponding 3D points for each landmark.
    \item \textbf{Reprojection and Error calculation.} We reproject these 3D points onto each image and compute the RE as the L2 distance between the original landmark and the reprojected 3D point normalized by image resolution, and average error across all images. 
\end{enumerate}
This approach evaluates multiview generation models by focusing on their ability to produce 3D-consistent results rather than adhering to a fixed ground truth. 
The \reprojectionerror provides a valuable basis for comparison across different methods. Furthermore, by computing RE on a set of real-world images that are independent of our generated images, we can establish a baseline that quantifies the error due to the noisy predictions from SuperGlue and SuperPoint rather than the quality of our generated images.

\heading{Naming Convention (RE@SG).} Please note that the use of SuperPoint~\cite{detone2018superpoint} and SuperGlue~\cite{sarlin2020superglue} estimation modules is a particular instantiation of our metric, and these could be replaced in the future with stronger counterparts such as MAST3R~\cite{dust3r_cvpr24,mast3r_arxiv24} or domain-specific keypoint detectors such as Sapiens~\cite{khirodkar2025sapiens}. Thus we use the naming convention of RE@SG to denote the Reprojection Error (RE) under SuperGlue (SG) estimation, which could be modified accordingly in the future for different estimators.

\section{Experiments}
\label{sec:experiments}

We present details of the datasets used in all training and validation stages, followed by discussion of evaluation metrics -- with particular emphasis on the 3D consistency metric and conclude with core experimental results and ablations.

\subsection{Data} 
\label{ssec:data}

\heading{\ig Dataset.}  %
We utilize a large proprietary dataset of approximately $3$ billion human-centric in-the-wild images for pretraining.
We provide more details on data filtering and curation in ~\cref{app_sec:data_filter}.

\vspace{1mm}
\heading{Head and Full-body Studio Datasets.} We rely on high-quality proprietary studio captures as our primary source of data for learning 3D consistency. 
Our model comes in two variants: head-only and full-body, each trained 
(for \midtraining and \posttraining stages) on corresponding datasets.  For our full-body model, we utilize a dataset of $861$ subjects ($811$ train, $50$ test), nearly $1000$ frames per subject.
For our head-only model, we use a dataset of $1450$ subjects ($1400$ train, $50$ test), nearly $40000$ frames per subject. The studio setup is similar to~\cite{martinez2024codec}, with two capture domes for capturing full-body and head-only high-resolution 4K images with $230$ and $160$ cameras.

\vspace{1mm}
\heading{\mobile Dataset.} To evaluate real-world performance, we collect casual images
of $50$ test subjects in an indoor office environment using an iPhone 13 Pro. We preprocess these images with Sapiens-2B~\cite{khirodkar2025sapiens} for
background segmentation before model inference. This dataset serves exclusively for evaluating our model's performance on in-the-wild inputs.

\subsection{Evaluation Setup and Metrics}

\heading{3D Consistency.} 
Following prior works, we report standard metrics like PSNR, SSIM ($\times 100$), and LPIPS ($\times 100$) metrics. However, these metrics unfairly penalize plausible novel views generated under incomplete inputs. We therefore introduce the \reprojectionerror metric, described in ~\cref{ssec:consistencymetric}, which validates 3D consistency without directly relying on ground truth. Our evaluation generates 4 randomly selected views from the test split.

\vspace{1mm}
\heading{Identity Preservation.} We use two metrics that measure identity preservation across generated views by computing cosine distance between features extracted via FaceNet~\cite{schroff2015facenet, vggface} for face similarity; and CLIP~\cite{radford2021learning} vision encoder for full-body similarity.

\vspace{1mm}
\heading{Pretrained Model Evaluation.} We measure the effectiveness of our pre-training strategy by reporting the 
FID~\cite{FID}. We use a smaller annotated 30M subset of \ig for training image and text-conditioned \pre@128 models, 
a 30M subset of unfiltered dataset for training No Filtering \pre@128, and a 1000 sample test set from \mobile dataset.

\subsection{Results} \label{ssec:results}

\heading{Pretraining and Data Filtering.} \cref{tab:pretrain} presents our pre-trained model results (Row 1) and ablations on Human-centric data filtering and Image-conditioned pretraining (Rows 2-5). Human-centric filtering and image-based conditioning are both critical to achieve high-quality generation.
The qualitative results are shown in \cref{app_sec:pretrain}.

\begin{table}[h!]
    \centering
    \small
    \setlength{\tabcolsep}{5pt}
    \begin{tabular}{llccccc}
        \toprule
         {\#} & \textbf{Method (Stage @ Resolution)} & FID $\downarrow$ \\
        \midrule
         {1.} & \ourmodel (\pre@512) & \textbf{51.164} \\
        \midrule
         {2.} & Human-centric Filtering (\pre@128)  & \textbf{75.639} \\
         {3.} & No Filtering (\pre@128)  & 86.838 \\
         \midrule
         {4.} & Image-conditioned (\pre@128)  & \textbf{75.639} \\
         {5.} & Text-conditioned (\pre@128) & 109.720 \\
        \bottomrule
    \end{tabular}
    \caption{
        \textbf{Pretraining and Data Filtering.} We report results of the full pretrained model \pre@512, and compare
        several variants of \pre@128 models. We report FID on \mobile dataset (1k samples).
    }
    \label{tab:pretrain}
    \vspace{-10pt}
\end{table}

\begin{figure*}[h!]
    \centering    
    \includegraphics[width=\textwidth]{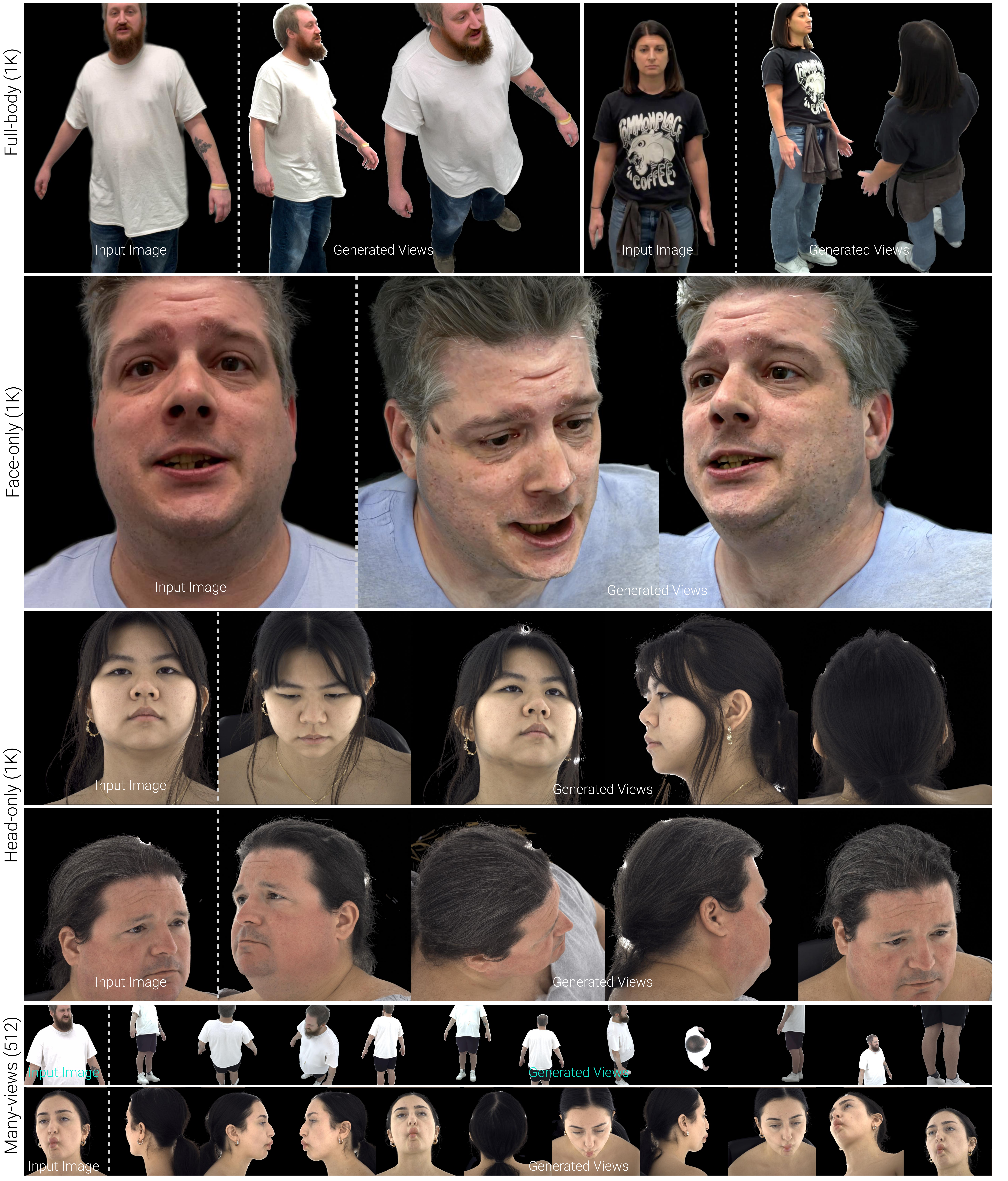}
    \captionof{figure}{\textbf{High Resolution Multi-view Generation of Unseen subjects.} \ourmodel enables generation of high-resolution 1K images given only a single image as input (left most in each block, separated with a dashed line). First row LHS shows generation from mobile captured photo, while the RHS shows unseen studio subject. Second row shows mobile captured face-only generations. Third and fourth row shows unseen studio subjects. Last two rows demonstrate simultaneous generation of 10 novel views given unseen studio subject.}
    \label{fig:qual}
    \vspace{10pt}
\end{figure*}

\vspace{1mm}
\heading{High-resolution Multi-view Generation.} In ~\cref{tab:result}, we evaluate 3D reconstruction and identity preservation for unseen subjects from the studio datasets. 
We show that increasing the output resolution of generation in our approach does not hurt 3D consistency or similarity. 
We put corresponding visuals in ~\cref{fig:qual}, Rows 2 and 3.
\begin{table}[h!]
    \centering
    \small
    \setlength{\tabcolsep}{2pt}
    \begin{tabular}{cllcccccc}
        \toprule
        & & & \multicolumn{4}{c}{\textbf{3D Consistency}} & \multicolumn{2}{c}{\textbf{Similarity}}  \\
        \cmidrule(r){4-7} \cmidrule(lr){8-9}

         & {\#} & \textbf{Split \& Resolution} & RE@SG $\downarrow$ & PSNR $\uparrow$ & SSIM $\uparrow$ & LPIPS $\downarrow$ & Face $\uparrow$ & Body $\uparrow$ \\

        \midrule
        \multirow{3}{*}{\STAB{\rotatebox[origin=c]{90}{{\textbf{\tiny \upperbody}}}}}

         & {1.} & Studio  (128)  & 3.3\textsubscript{\color{red}{+0.8}} & 21.0 & 67.6 & 14.8 & {62.0}\textsubscript{{\color{red}{-13.3}}} & {61.5}\textsubscript{{\color{red}{-16.2}}} \\

         & {2.} & Studio (1K)  & 3.4\textsubscript{\color{red}{+0.5}} & 20.3 & 72.0 & 26.2 & {73.5}\textsubscript{{\color{red}{-2.5}}} & {79.4}\textsubscript{{\color{red}{-0.1}}} \\

         & {3.} & \mobile Face (1K)  & 3.0 & - & - & - & {67.6} & - \\

        \midrule

        \multirow{3}{*}{\STAB{\rotatebox[origin=c]{90}{{\textbf{\tiny \fullbody}}}}} 
         & {4.} & Studio  (128)  & 3.6\textsubscript{\color{red}{+0.1}} & 22.8 & 84.0 & 10.0 & {41.7}\textsubscript{{\color{red}{-8.5}}} & {67.1}\textsubscript{{\color{red}{-9.4}}} \\
         & {5.} & Studio  (1K)  & 1.5\textsubscript{\color{red}{+0.1}} & 22.4 & 91.7 & 11.1 & {64.7}\textsubscript{{\color{red}{-6.0}}} & {74.1}\textsubscript{{\color{red}{-2.2}}} \\

         & {6.} & \mobile  (1K)  & 1.7 & - & - & - & {58.0} & {68.1} \\

        \bottomrule
    \end{tabular}
    \caption{
        \textbf{Results on Unseen Studio and \mobile data (\cref{ssec:results}.}) We report results on Post-trained \ourmodel models at three different resolutions. We report 3D metrics PSNR, SSIM, and LPIPS; as well as our proposed Reprojection Error (RE@SG) under SuperGlue estimation which does not require ground truth (\cref{ssec:consistencymetric}). We report 2D Face and Body similarities. \textcolor{red}{Red} subscript show deviation against ground truth value.}
    \label{tab:result}
    \vspace{-10pt}
\end{table}

\vspace{1mm}
\heading{Generations from Casual \mobile Photos.} \noindent We present in ~\cref{tab:result} (Rows 3,6) \reprojectionerror and similarity scores for casually taken images from the \mobile dataset with 1K resolution model. In this scenario, the standard reconstruction error metrics cannot be assessed due to missing ground truth. We find that the reprojection error on \mobile captures remains comparable to the studio dataset -- demonstrating 3D consistency. This illustrates the generalizability of Pippo beyond the multi-view training data domain, where our pretraining with large-scale in-the-wild human data is critical.
We put corresponding visuals in ~\cref{fig:qual}, Row 1.

\begin{figure*}
    \includegraphics[width=\linewidth]{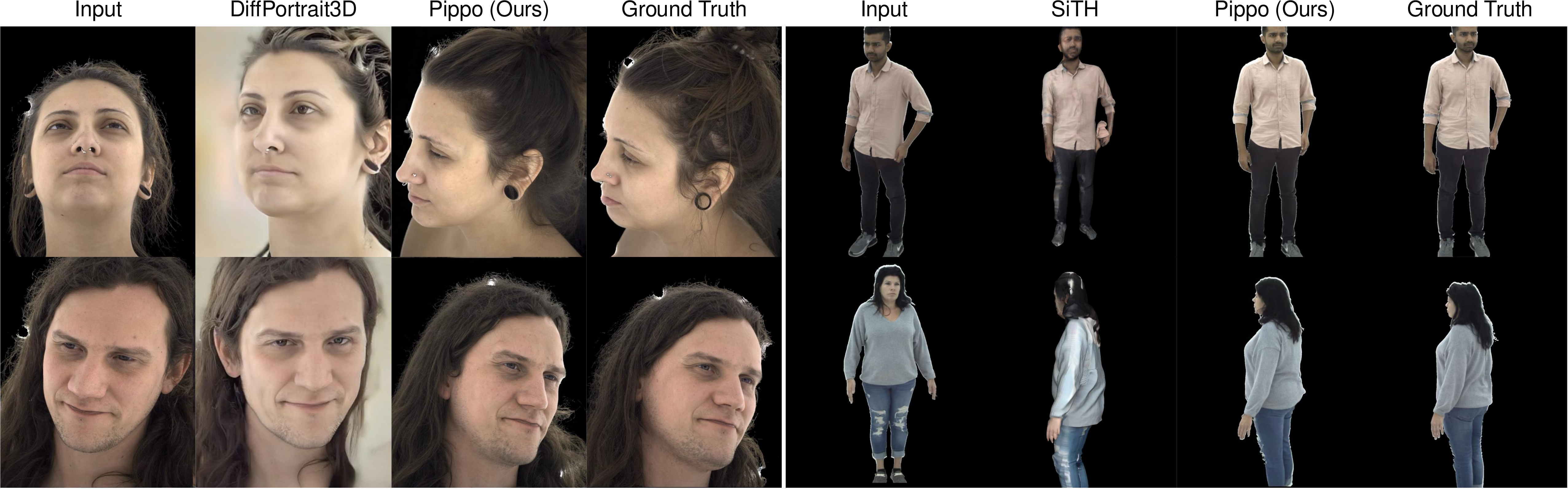}
    \vspace{-0.2in}
    \captionof{figure}{\textbf{Visual comparison with state-of-the-art methods}. We visually compare \ourmodel with state-of-the-art baselines DiffPortrait3D~\cite{gu2024diffportrait3d} (head-only) and SiTH~\cite{ho2024sith} (full-body) generation.} 
    \label{fig:baseline}
\end{figure*}

\heading{Comparisons with External Benchmarks.} In ~\cref{fig:baseline}, we compare \ourmodel with individual state-of-the-art baselines in \fullbody and \upperbody generation. SiTH~\cite{ho2024sith} reconstructs textured human-mesh using ControlNet paired with a SDF representation. Compared to SiTH, \ourmodel facilitates high-resolution and accurate multiview synthesis. DiffPortrait3D~\cite{gu2024diffportrait3d} inverts a 3D-GAN based on a given input image. Compared to it, our model supports greater viewpoint variability and ensures closer adherence to the input image.

\heading{Quantitative comparisons and baselines.} Existing SoTA Human methods~\cite{ho2024sith, gu2024diffportrait3d} use explicit SMPL priors, and thus are difficult to compare with directly. Qualitatively, we found that they cannot handle novel views or preserve details (~\cref{fig:baseline}), and hence do not quantitatively compare against them. 
\noindent In Pippo, we focus on creating a strong multi-view human generator, and we benchmark four state-of-the-art multi-view diffusion models on the {iPhone full-body dataset} in~\cref{tab:baseline}. We find that Pippo preserves identity (\ie face and body similarity) and 3D consistency (RE) better while also operating at a higher resolution compared to baselines.

\begin{table}[h!]
    \centering
    \small
    \setlength{\tabcolsep}{3pt}
    \begin{tabular}{c l c cc}
        \toprule
         {\#} & \textbf{Method \& Resolution} & RE@SG $\downarrow$ & Face $\uparrow$ & Body $\uparrow$ \\
        \midrule
         {1.} & MV-Adapter~\cite{huang2024mvadapter} (768)  & 4.7 & \underline{43.0} & {64.1} \\
         {2.} & Era3D~\cite{li2024era3dhighresolutionmultiviewdiffusion} (512)  & \underline{4.1}  & {38.1} & \underline{64.4} \\
         {3.} & Wonder3D~\cite{long2023wonder3d} (256)  & 5.3  & {34.7} & {58.8} \\
         {4.} & \ourmodel (P3@1K)  & \textbf{3.0} & \textbf{58.0} & \textbf{68.1} \\
        \bottomrule
    \end{tabular}
    \caption{\textbf{Quantitative comparison with SoTA multi-view models.} We quantitatively compare \ourmodel against state-of-the-art multi-view diffusion models. We find that Pippo preserves identity (\ie face and body similarity) and 3D consistency (RE) better while operating at a higher resolution compared to baselines.}
    \label{tab:baseline}
\end{table}

\heading{Benchmarking \ourmodel on public datasets.} In~\cref{tab:public}, we benchmark \ourmodel on the public Codec Avatar datasets~\cite{martinez2024codec} to aid future comparisons. Specifically, we use Ava-256 containing 256 head-only captures and Goliath containing 4 full-body. During evaluation, we only use subsets of these datasets that were not used for training. We find that \ourmodel's performance on these datasets is inline with its performance on internal studio datasets. 

\begin{table}[h!]
    \centering
    \small
    \setlength{\tabcolsep}{1pt}
    \begin{tabular}{llcccccc}
    \toprule
     {\#} & \textbf{Dataset (P3@1K)} & RE@SG $\downarrow$ & PSNR $\uparrow$ & SSIM $\uparrow$ & LPIPS $\downarrow$  & Face $\uparrow$ & Body $\uparrow$ \\

        \midrule

        {1.} & Ava-256 (\upperbody) &  3.8 & 20.1 & 68.7 & 26.6 & {72.9} & {76.8} \\
        {2.} & Goliath (\fullbody)  &   1.2 & 20.4 & 89.7 & 15.5 & {87.5} & {77.7} \\

        \bottomrule
    \end{tabular}
    \caption{\textbf{Benchmarking \ourmodel on public Ava-256 and Goliath datasets.} We find that \ourmodel's performance on these datasets is inline with its performance on internal studio datasets. This will aid future comparisons against \ourmodel.}
    \label{tab:public}

\end{table}

\subsection{Ablations}

We examine design choices at each training stage (\cref{sec:method}) and present our ablation results in \cref{tab:ablation}. Experiments are conducted at $128\times128$ resolution on the \upperbody dataset.

\begin{table}[h!]
    \centering
    \small
    \setlength{\tabcolsep}{1pt}
    \begin{tabular}{llcccccc}
        \toprule
        & & \multicolumn{4}{c}{\textbf{3D Consistency}} & \multicolumn{2}{c}{\textbf{Similarity}}  \\
        \cmidrule(r){3-6} \cmidrule(lr){7-8}

         {\#} & \textbf{Method (Stage @ Res.)} & RE@SG $\downarrow$ & PSNR $\uparrow$ & SSIM $\uparrow$ & LPIPS $\downarrow$  & Face $\uparrow$ & Body $\uparrow$ \\

        \midrule

        {1.} & \ourmodel (\post@512) & 2.7\textsubscript{\color{red}{+0.7}} & 21.7 & 71.7 & 21.5 & {74.1}\textsubscript{{\color{red}{-2.0}}} & {77.4}\textsubscript{{\color{red}{-2.4}}} \\

        {2.} & \ourmodel Face-only  (\post@512) & 2.4\textsubscript{\color{red}{+0.4}} & 20.5 & 70.3 & 25.3 & {74.3}\textsubscript{{\color{red}{-1.7}}} & {75.8}\textsubscript{{\color{red}{-4.0}}}  \\

        {3.} & No Mid-train (\post@512) & 5.6\textsubscript{\color{red}{+3.6}} & 14.5 & 59.5 & 44.2 & {20.4}\textsubscript{{\color{red}{-55.6}}} & {63.4}\textsubscript{{\color{red}{-16.4}}} \\

        \midrule

        {4.} & \ourmodel (\post@128)   & 3.3\textsubscript{\color{red}{+0.8}} & 21.0 & 67.6 & 14.8 & {62.0}\textsubscript{{\color{red}{-13.3}}} & {61.5}\textsubscript{{\color{red}{-16.2}}} \\

        {5.} & No Anchor (\post@128) & 11.5\textsubscript{\color{red}{+9.0}} & 17.1 & 54.0 & 21.1 & {63.1}\textsubscript{{\color{red}{-12.2}}} & {68.3}\textsubscript{{\color{red}{-9.3}}} \\

        {6.} & No \plucker (\post@128) & 4.2\textsubscript{\color{red}{+1.7}} & 20.2 & 64.5 & 16.1 & {63.5}\textsubscript{{\color{red}{-11.8}}} & {66.1}\textsubscript{{\color{red}{-11.5}}} \\

        \midrule

        {7.} & \ourmodel (\mid@128) & 3.4\textsubscript{\color{red}{+0.9}} & 19.1 & 61.4 & 17.2 & {60.0}\textsubscript{{\color{red}{-15.3}}} & {62.6}\textsubscript{{\color{red}{-15.0}}}  \\

        {8.} & No Pre-train (\mid@128) &  5.9\textsubscript{\color{red}{+3.4}} & 13.1 & 39.3 & 44.9 & {19.5}\textsubscript{{\color{red}{-55.8}}} & {49.0}\textsubscript{{\color{red}{-28.6}}} \\

        {9.} & Cross-Attn (\mid@128) & 3.5\textsubscript{\color{red}{+1.0}} & 18.0 & 59.2 & 22.1 & {66.2}\textsubscript{{\color{red}{-9.1}}} & {70.7}\textsubscript{{\color{red}{-7.0}}} \\

        {10.} & Non-frontal (\mid@128)  & 5.8\textsubscript{\color{red}{+3.3}} & 15.2 & 52.4 & 30.1 & {49.2}\textsubscript{{\color{red}{-26.1}}} & {60.7}\textsubscript{{\color{red}{-16.9}}} \\

        \bottomrule
    \end{tabular}

    \caption{
        \textbf{Ablation on design choices and training stages.} 
        We evaluate several multi-view models at \resone and \resthree resolution at different training stages on \upperbody dataset. We ablate the choice of doing Mid-training and Pre-training, along with different spatial controls and reference conditioning methods..The training stages are labeled as \mid \ (\midtraining) and \post \ (\posttraining), followed by @ specified resolution.}
    \label{tab:ablation}

\end{table}

\heading{Significance of Pre-training and Mid-training.} Pre-training our model on the \ig dataset enables robust generalization to novel identities, as demonstrated in \cref{tab:ablation}, Row 8. Without pre-training, model generalization deteriorates, resulting in unclear facial features. Skipping mid-training at lower resolution impairs consistent multi-view generation, as shown in Row 2.

\heading{Importance of Frontal Input Reference.} Our ablation in \cref{tab:ablation}, Row 10 demonstrates that completely randomizing the view point of an input reference image leads to overfitting to training identities. Non-frontal views, especially rear views, have very limited information about the identity which forces the network to pick up spurious correlations.

\heading{Importance of Self-attention.} Replacing self-attention with cross-attention for reference image encoding, using the same routing as image-conditioned pretraining from \cref{ssec:base_model}, leads to degraded performance as shown in \cref{tab:ablation}, Row 9.
We observe that this setup causes the model to ignore input conditioning, generating images that only vaguely resemble training subjects.

\heading{Role of large scale Humans-3B pre-training dataset.} We utilized intermediate checkpoints from pre-training stage (\pre) trained on 30\% and 70\% of the data, and a separate checkpoint trained on 1\% high-quality subset of Humans-3B. 
Starting from these checkpoints, we mid-trained \ourmodel to denoise 4 views at \resone resolution on \fullbody dataset for two days. We report their respective results in~\cref{tab:dataset_ablation}. 
We found large-scale data is crucial for generalization to novel identities -- indicated by high gains in face similarity metric.

\begin{table}[h!]
    \centering
    \small
    \setlength{\tabcolsep}{1pt}
    \begin{tabular}{llcccccc}
        \toprule

         {\#} & \textbf{Pretrain Data (\mid@128)} & RE@SG $\downarrow$ & PSNR $\uparrow$ & SSIM $\uparrow$ & LPIPS $\downarrow$  & Face $\uparrow$ & Body $\uparrow$ \\

        \midrule

        {1.} & 30M (1\% HQ) &  \underline{3.8} & 21.3 & 81.8 & 13.3 & {30.6} & {66.2} \\

        {2.} & 900M (30\% Random)   &   4.0 & 21.6 & 82.0 & \textbf{12.1} & {30.6} & \underline{67.2} \\

        {3.} & 2.1B (70\% Random) &   \textbf{3.7} & 21.6 & 82.1 & \textbf{12.1} & \underline{37.6} & \textbf{67.5} \\

        {4.} & Humans-3B (100\%) &  \textbf{3.7} & \textbf{21.9} & \textbf{82.3} & \underline{12.7} & \textbf{41.7} & \underline{67.2} \\

        \bottomrule
    \end{tabular}
    \vspace{-8pt}
    \caption{\textbf{Ablating the size of pre-training dataset on Humans-3B.} We conduct \fullbody mid-training with pretrained checkpoints trained on 1\%, 30\%, 70\%, and 100\% subsets of Humans-3B dataset. We found large-scale data is crucial for generalization to novel identities (\ie face similarity).}
    \label{tab:dataset_ablation}

\end{table}

\begin{figure*}[h!]
    \centering    
    \includegraphics[width=\textwidth]{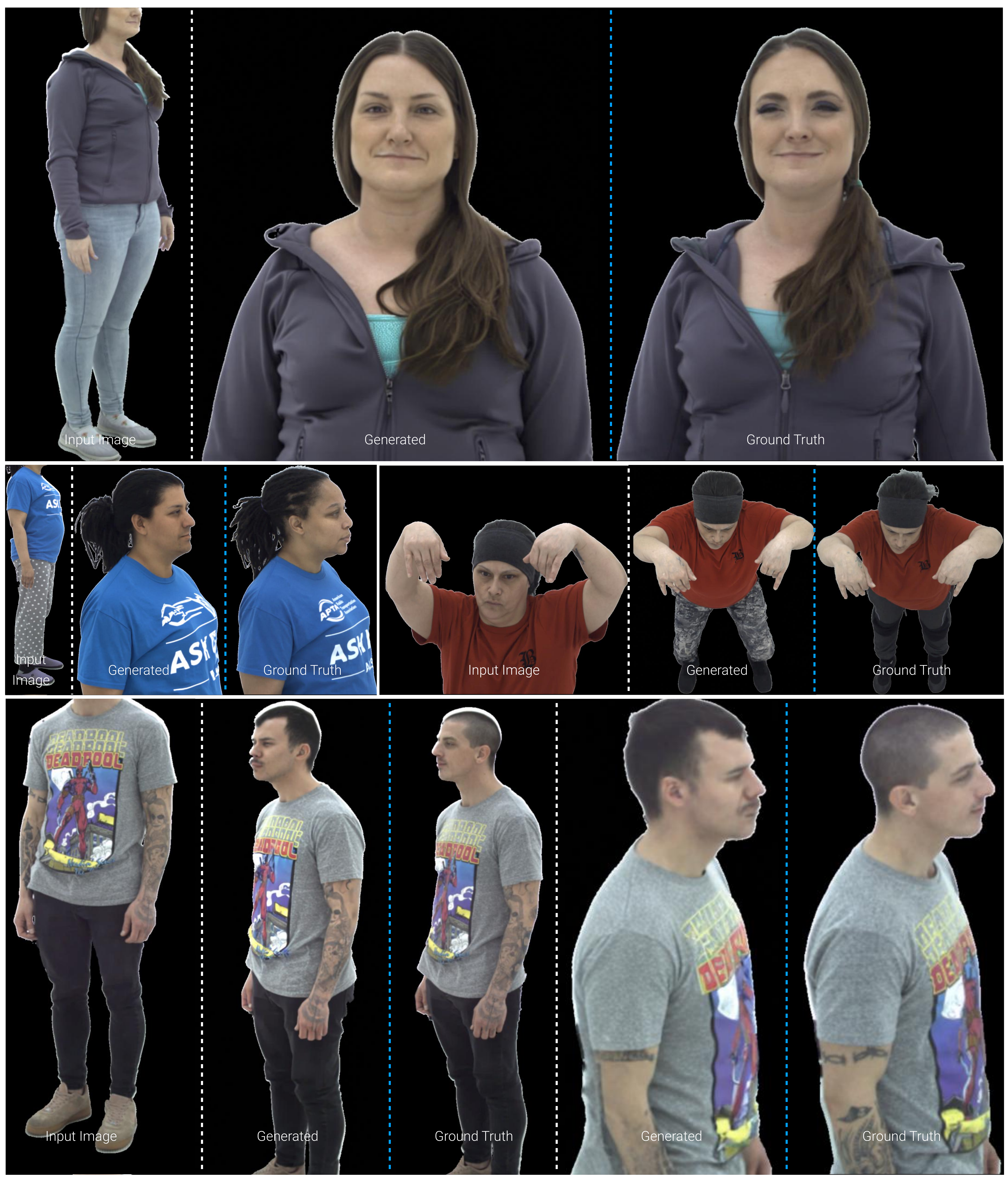}
    \captionof{figure}{\textbf{\ourmodel can handle occluded inputs.} We show \ourmodel's generations given incomplete input images -- such as partially or fully occluded faces, unobserved t-shirt designs on test split of \fullbody dataset. We show corresponding ground truth separated with blue dotted line -- it can be seen that \ourmodel faithfully follows the known content while auto-completing unseen segments.}
    \label{fig:qual_incomplete}
    \vspace{15pt}
\end{figure*}

\section{Conclusion}
We present Pippo, a diffusion transformer model that generates a dense set of high-resolution 
multi-view consistent images of a person from a single image. 
Our experiments show that our multi-stage training strategy, which combines large-scale 
in-the-wild data with high-quality multi-view studio captures, enables generalizable 
high-resolution multi-view synthesis.
Analysis of the diffusion transformer architecture reveals that self-attention with the reference image, 
\plucker coordinates with SIREN, and Spatial Anchor are all essential for high-fidelity multi-view human 
generation.
Pippo achieves, for the first time, consistent multi-view human image generation at 1K resolution. 
We also show that our proposed 3D consistency metric enables evaluation of 3D consistency without 
paired ground-truth data. 
One of the limitations of our approach is the limited number of simultaneously generated views, 
originating from large context length and memory constraints, which can be tackled with parallelization 
techniques and autoregressive generation.
Extending our approach to multi-view consistent video generation will be addressed in future work.

\clearpage
\newpage
\bibliographystyle{assets/plainnat}
\bibliography{main}

\clearpage
\newpage
\beginappendix

\section{DiT Architecture and Training}\label{app_sec:dit}

\heading{Architecture.} We use a Diffusion Transfomer with $28$ DiT + ControlMLP blocks (depth) in \ourmodel operating at $1536$ dimension. The DiT blocks account for $1.3$B parameters and ControlMLP blocks for $248$M parameters respectively. We use 
latent autoencoder which provides $8$x spatial downsampling, and $4$-channel latent space. 
It contains $101$M parameters. We find the placement of the ControlMLP block modulation in DiT to be crucial, specifically, being placed \textit{before} the scale, shift, and gate is important for the control signal to work well. 

\heading{Training Stages.} We pre-train (P1@256) our model on $512$ A100 GPUs for $1$M iterations, with batch size of $24$ per GPU and at \restwo resolution. 1M on base model, and 700K iterations on 512x512 resolution. During pretraining \ourmodel, we use image-only conditioning via 
DinoV2~\cite{oquab2023dinov2} using ViT-L/14 based models to provide weak supervision and alignment for target generations. We mid-train our model on \fullbody and \upperbody datasets at \resone resolution for nearly $100$K steps, and post-train our models with spatially-aligned conditions at \resthree and \resfour for nearly $50$K steps. We pre-train, mid-train, and post-train for roughly 2:1:1 weeks, respectively.

\heading{Optimizer and Hyperparameters.} We use LR of $1e$-4 during all stages and AdamW optimizer with beta values set to $[0.9, 0.98]$ and epsilon to $1e$-6. We use linear warmup for initial 1000 steps starting at $1e$-6. We conduct training at full precision (float32), and run inference at half precision (bfloat16).

\section{Diffusion Model and Inference Settings}
\label{app_sec:diffusion}

We use classifier-free guidance(CFG)~\cite{ho2022classifierfree} during training, with $20$\% dropout probability on input reference image. We use 50 DDIM steps for sampling. During inference, we find that lower CFG scales between $3-5$ work better at higher resolution and generating viewpoints that are closer to input view, this observation is inline with Stable Video Diffusion~\cite{blattmann2023stable} which proposed to linearly increase CFG scale after the first frame generation. Inference speed with \ourmodel depends on the resolution and number of views to be denoised; we report few combinations in ~\cref{tab:speed}.

\begin{table}[h!]
    \centering
    \small
    \setlength{\tabcolsep}{10pt}
    \begin{tabular}{llccccc}
        \toprule
         {\#} & \textbf{Method (Stage @ Resolution)} & Views & Speed (sec) $\downarrow$ \\
        \midrule
         {1.} & Pretrained (\pre@256) & 1 & 2.51  \\
         {2.} & Pretrained (\pre@512) & 1 & 2.59 \\
        \midrule
         {3.} & Mid-trained (\mid@128)  & 4 & 6 \\
         {4.} & Mid-trained (\mid@128)  & 48 & 14 \\
         \midrule
         {5.} & Post-trained (\post@512)  & 4 & 40 \\
         {6.} & Post-trained (\post@512)  & 48 & 490\\
         \midrule
         {7.} & Post-trained (\post@1K)  & 4 & 185 \\
         {8.} & Post-trained (\post@1K)  & 12 & 622 \\
        \bottomrule
    \end{tabular}
    \caption{
        \textbf{Inference Speed of \ourmodel.} We show inference speed without any optimizations (using bfloat16) against varying resolution and number of views being generated.
    }
    \label{tab:speed}
    \vspace{-10pt}
\end{table}

\begin{figure}[h!]
    \centering    
    \hspace{-0.5cm}
    \includegraphics[width=\linewidth]{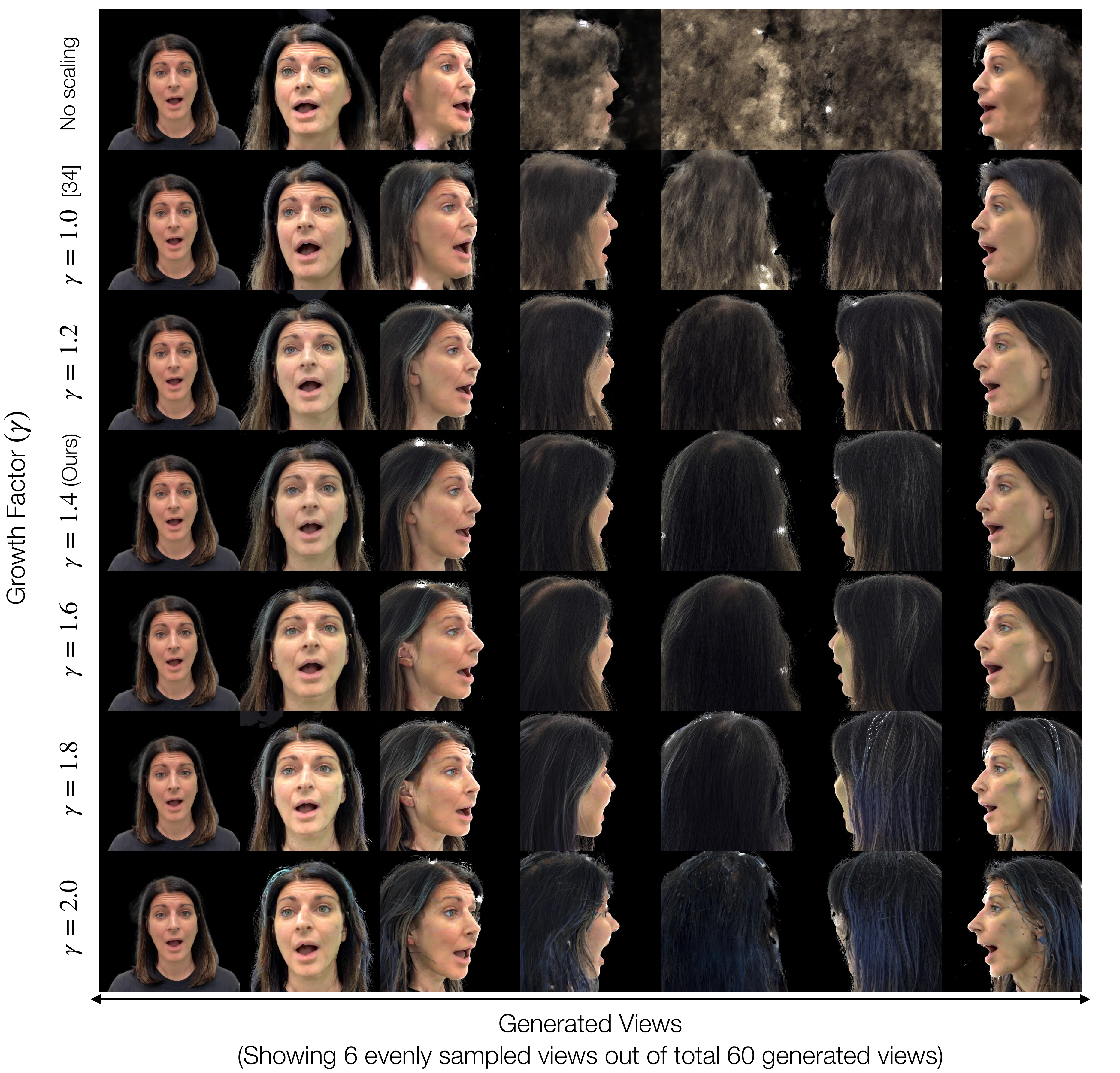}
    \vspace{-0.2cm}
    \caption{{\textbf{Generations under varying strengths of growth factor $\gamma$ (\cref{ssec:scaling_views})}. On each row we show the generated views across vanilla attention~\cite{vaswani2017attention} (No scaling), prior work~\cite{jin2023training} and our formulation Eq.~\eqref{attn_bias}. Growth factor ($\gamma$) greater than 1.0 helps mitigate the entropy buildup, however increasing $\gamma$ beyond $1.6$ leads to oversaturation artifacts (somewhat akin to high CFG scale). We show only 6 views per row subsampled evenly from 60 views generated at \resthree resolution. The model was trained to jointly denoised only 12 views ($N_i = 5*N_t$).}}
    \label{fig:qual_entropy_full}
\end{figure}

\heading{Attention Biasing.} To compute the entropy (shown in ~\cref{fig:entropy}), we use the first, middle, and last blocks of Pippo; and aggregate over all attention heads, conditional and unconditional inference passes, and DDIM steps. In~\cref{fig:qual_entropy_full}, we show visuals for increasing strength of the scaling growth factor ($\gamma$) when generating 60 views simultaneously at 512x512 resolution. It is evident that using this attention biasing is quite crucial in making diffusion models generalize across many views (long context sequences). Growth factor ($\gamma$) greater than 1.0 helps mitigate the entropy buildup; however, increasing $\gamma$ beyond $1.6$ leads to over-saturation artifacts somewhat akin to ones caused by high CFG scale. 

\heading{Varying Classifier-free Guidance (CFG) using a bump function.} 
Classifier-free Guidance guidance enables a trade-off between diversity and realism~\cite{ho2022classifierfree}; with higher values of CFG resulting in diverse generations (\ie higher Inception Score) and lower values of CFG leading to phtorealistic generations (\ie higher FID)
In our setup, the single image to multi-view task involves faithfully preserving known content while also hallucinating diverse possibilities of unseen regions. The amount of known content and unknown content varies for each view that is being generated; for example, the back of the heads are often unseen; however, central parts of the face, such as the nose, are generally specified in the frontal input image. 
We can use this information to rescale the CFG weight for each view separately.
Specifically, we find that using a lower weight for regions where content copying is required prevents saturation artifacts, whereas increasing the CFG scale of unseen regions leads to more stable and diverse generations. 
Thus, we increase the CFG linearly starting from the front facing view at $0\degree$ azimuth until $90\degree$ azimuth (side-view) where it reaches its peak value. Then we keep the CFG scale fixed at this peak value until the azimuth reaches $270\degree$ (opposite side-view), and finally decrease it linearly back to starting value at azimuth of $360\degree$. This is a bump function and we find that starting with CFG scale between $[7.0, 9.0]$, and having a peak CFG scale between $[15.0,19.0]$ results in reduced artifacts and diverse generations especially in the unseen regions. A similar trick is also used in image-to-video work of SVD~\cite{blattmann2023stable}. 

{
\heading{Rescaling diffusion timesteps under varying resolution.} Prior works~\cite{chen2023importance, esser2024scaling} suggest that as resolution increases the noise scale has to be shifted to ensure same level of corruption. Based on this Stable Diffusion 3~\cite{esser2024scaling} derive a noise reweighing scheme by assuming degradation to a constant-pixel image and ensuring that the uncertainity under degradation for each pixel stays constant. Since SD3 uses conditional flow-matching objective and we train Pippo using the DDPM~\cite{DDPM} objective, we cannot use their reweighing scheme directly. Here, we provide a derivation for an equivalent reweighing scheme for DDPM objective. We can define forward process as:
   \[
   z_t = \sqrt{\alpha_t} z_0 + \sqrt{1 - \alpha_t} \epsilon,
   \]
where \(\alpha_t\) is a monotonically decreasing function of \(t\). The uncertainty in DDPM is governed by the variance of the forward process, which depends on \(\alpha_t\) and \(1 - \alpha_t\). Consider a constant image \(z_0 = c \mathbbm{1}\), where \(c \in \mathbb{R}\) and \(\mathbbm{1} \in \mathbb{R}^n\). The forward process in DDPM produces:
\[
z_t = \sqrt{\alpha_t} c \mathbbm{1} + \sqrt{1 - \alpha_t} \epsilon,
\]
where \(\epsilon \sim \mathcal{N}(0, I)\). The uncertainty in estimating a constant-valued image \(c\) is, where $n$ are total number of pixels in the image:
\[
\sigma(t, n) = \frac{\sqrt{1 - \alpha_t}}{\sqrt{\alpha_t}} \cdot \frac{1}{\sqrt{n}}.
\]

To map a timestep \(t_n\) at resolution \(n\) to a timestep \(t_m\) at resolution \(m\) such that the uncertainty \(\sigma(t_n, n) = \sigma(t_m, m)\), we solve:
\[
\frac{\sqrt{1 - \alpha_{t_n}}}{\sqrt{\alpha_{t_n}}} \cdot \frac{1}{\sqrt{n}} = \frac{\sqrt{1 - \alpha_{t_m}}}{\sqrt{\alpha_{t_m}}} \cdot \frac{1}{\sqrt{m}}.
\]
Rearranging, we get the resolution-dependent timestep mapping for DDPM isolates \(\alpha_{t_m}\) as:
\[
\alpha_{t_m} = \frac{\alpha_{t_n}}{\frac{m}{n} + \alpha_{t_n} \left(1 - \frac{m}{n}\right)}.
\]
We use the above reweighing to rescale noise steps when training models at a higher resolution of \resthree and \resfour during post-training. In practice, we set $m/n$ ratio to be slightly lower than the actual value following SD3~\cite{esser2024scaling}.

\section{\ig: Filtering and Stats}\label{app_sec:data_filter}

We run image metadata filtering to keep images whose short edges are at least 720 pixels and file sizes are at least 120 KB.
We run Detectron2~\cite{wu2019detectron2} (pose detection) to keep images containing one clearly detected person (detection score at least 0.9 and the secondary clear person detection score is at most 0.4) with heads at least partially visible, and the long edge of the detected bounding box is at least 300 pixels.
We also use a custom person realism classifier to drop computer-generated or computer-processed imagery.
We provide rough statistics of our curated human-centric dataset in ~\cref{fig:data_stat}. We bucket these attributes in bins along X-axis and plot their respective sizes (normalized between $0-1$) on Y-axis. 

\begin{figure*}[h!]
    \centering    
    \includegraphics[width=\textwidth]{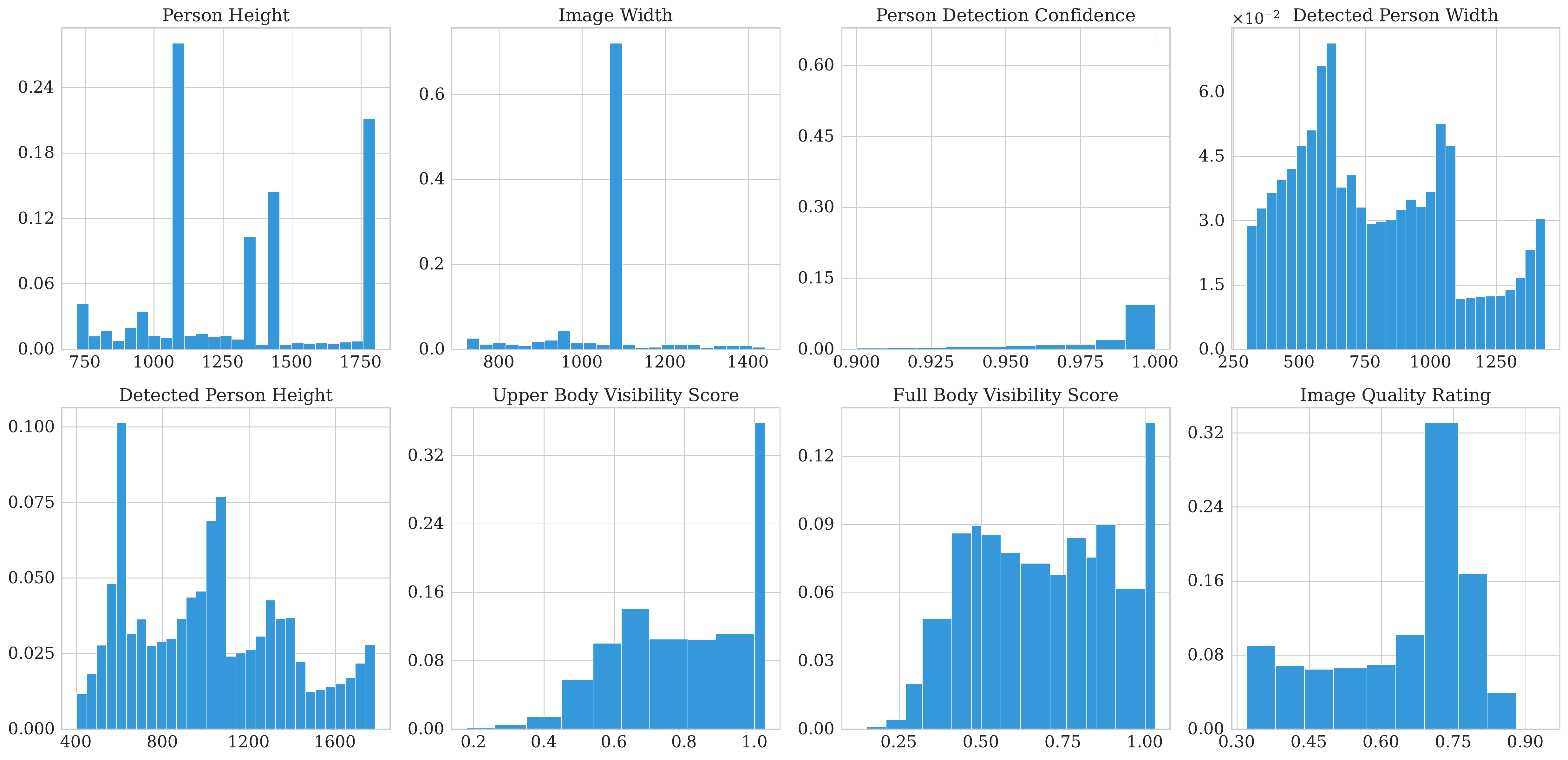}
    \caption{\textbf{Statistics of our curated \ig dataset}. We bucket these attributes in bins along X-axis and plot their respective sizes normalized between $[0,1]$ on Y-axis. Filtering with these statistics enable us to retain images with detected-person confidence and image quality.}
    \label{fig:data_stat}
\end{figure*}

\section{Webpage Visuals}

\heading{Webpage.} We upload a supplementary webpage which contains 360$\degree$ turnaround videos from our model generated for Full-body, Head-only and Face-only settings. Additionally, we put visuals where we provide as input a monocular video (framewise), and generate frames independently at each timestep. We find \ourmodel preserves the known details while hallucinating plausible unseen parts well. We also put the visualization of our spatial anchor and corresponding generations. See \texttt{\url{http://yashkant.github.io/pippo}} for the webpage.
\vspace{5pt}

\begin{figure*}[h!]
    \centering    
    \includegraphics[width=\textwidth,trim=0 10.5cm 0 10.25cm, clip]{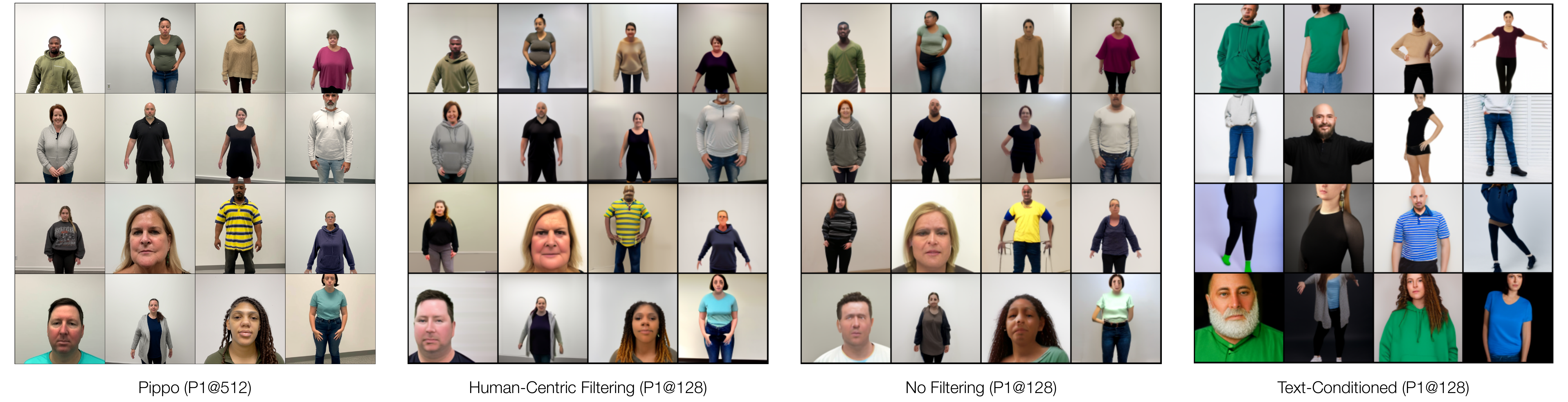}
    \caption{\textbf{Qualitiative and ablation visuals from pretrained model}. Consistent with quantitative 
    evaluation, visual quality of generated images improves with using human-centric filtering, and 
    image-conditioned models generate samples which are visually closer to the domain (casual \mobile captures). 
    There is also an obvious boost in quality due to higher resolution.}
    \label{fig:pretrain}
\end{figure*}

\section{Why the name Pippo?}
\begin{wrapfigure}[10]{r}{0.2\textwidth}
    \vspace{-30pt}
    \centering
    \includegraphics[width=\linewidth]{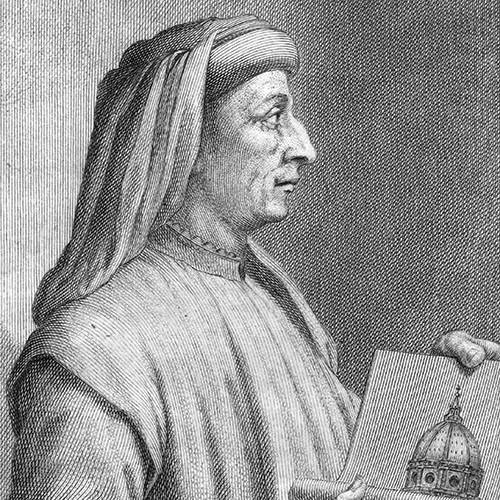}
    {\textbf{Filippo Brunelleschi.}}
    \label{fig:pippo}
\end{wrapfigure}

Our model is named after Filippo di ser Brunellesco di Lippo Lapi (1377 – 15 April 1446), widely recognized as Filippo Brunelleschi and affectionately known as Pippo by Leon Battista Alberti. 
Brunelleschi was an Italian architect, designer, goldsmith, and sculptor. 
He pioneered the application of vanishing points in artwork to achieve accurate perspective vision. Similarly, our model employs a single 3D spatial anchor to produce consistently improved images.

\section{Visuals from Pretrained Model (P1  )}\label{app_sec:pretrain}

In ~\cref{fig:pretrain}, we showcase qualitative visuals related to the findings in Table 2. It is evident that the model trained with filtered data and using an image-conditioned objective produces high-quality human figures.

\section{Frequently Asked Questions}

\heading{Ablation with Missing or Inconsistent spatial anchor.} 
Spatial Anchor acts as a placement signal for Pippo since we do not provide it with intrinsics or extrinsics of the input image. In~\cref{fig:anchor_ablation}, we run inference on the \upperbody \pre@512 model with missing spatial anchor, or when it is inconsistent with input head pose (rotated downwards at the floor by 90$\degree$). When the spatial anchor is missing, Pippo often generates an empty image because in our training data it implies that the subject's head is not visible in the generated view. We find that Pippo is robust to anchor rotations; this suggests that \ourmodel relies on the spatial anchor only for placement control and infers head-pose from the input image. 

\begin{figure}[h!]
    \centering    
    \includegraphics[width=\linewidth]{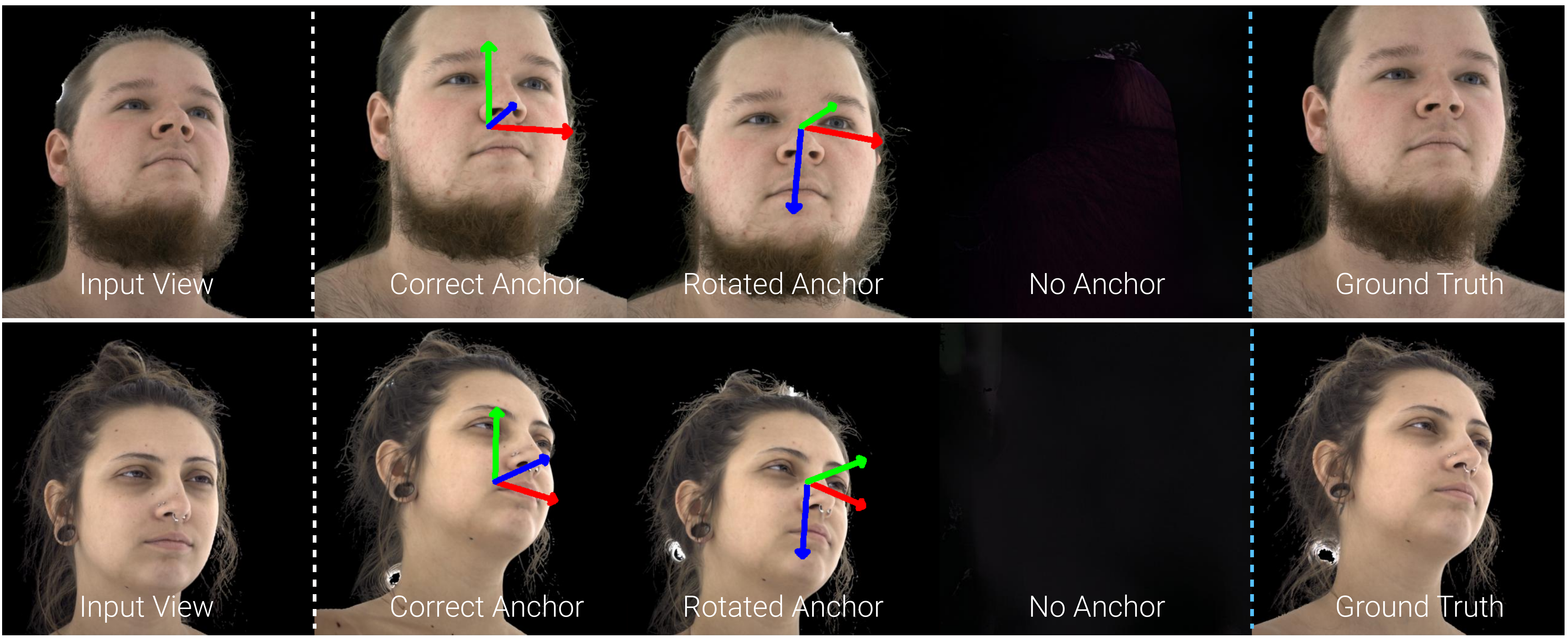}
    \caption{\textbf{Ablation with Missing or Inconsistent Spatial Anchors.} We run inference on the \upperbody \pre@512 model with missing spatial anchor, or when it is inconsistent with input head pose rotated downwards towards the floor by 90$\degree$.}
    \label{fig:anchor_ablation}
\end{figure}

\noindent\textbf{Can we reconstruct \ourmodel outputs?} 
Yes, we can reconstruct \ourmodel's generations directly using a NeRF or Gaussian Splatting. 

\heading{Why use spatial-control in post-training only?}
We find that skipping pixel-aligned controls in mid-training helps us to train faster and allows training jointly on a greater number of views. Additionally, since we mid-train at the low resolution of \resone, we find that pixel-aligned control needs to be re-injected during post-training again.

\noindent\textbf{Is Reprojection Error (RE) pairwise, and can it handle occlusions?} Yes, we compute \textit{the mean} Reprojection Error over pairs of images. We divide the generated images randomly into non-overlapping pairs and compute pairwise RE. 
We re-project the triangulated 3D points back to each of the two images to compute the RE. 
We only use high-quality correspondence matches by setting a threshold of $>0.2$ in SuperGlue, and reject image pairs that have fewer than $<5$ matches detected. This filtering helps to avoid spurious correspondences in distant and occluded views. We will release the code for metric.

\heading{Trends in overfitting experiments (in~\cref{ssec:spatial_control}) may change under largescale training.} Empirically, we found that the ability to generalize to novel viewpoints of a single scene under overfitting is correlated with a greater ability to steer the camera viewpoint and place the subject precisely after post-training. For example, in Tab. 4, Row 5 we can see that removing the spatial anchor drops the 3D consistency of the post-trained model the most and is correlated with the overfitting result in Tab. 1, Row 6. The distribution of cameras in overfitting remains similar to that of full training. Thus, we use overfitting as a proxy to compare existing spatial control modules cheaply.

\begin{figure*}[t!]

    \centering    
    \includegraphics[width=\textwidth]{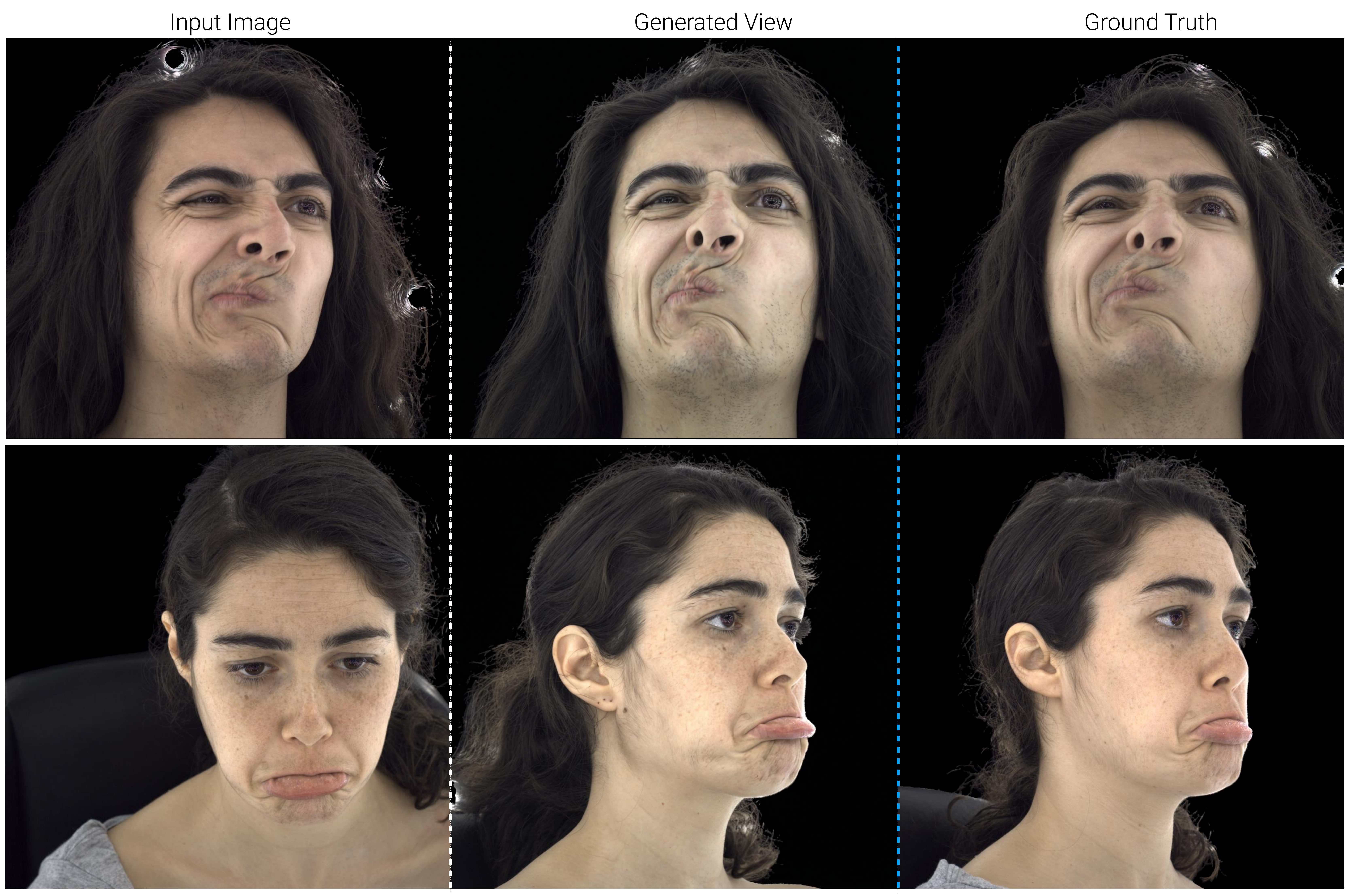}
    \caption{\textbf{\ourmodel can handle extreme facial expressions}. We show generations where reference image comes from unseen \upperbody subjects showing extreme facial expression alongside ground truth. We put similar visuals from our Face-only model on webpage.} 

    \includegraphics[width=\textwidth]{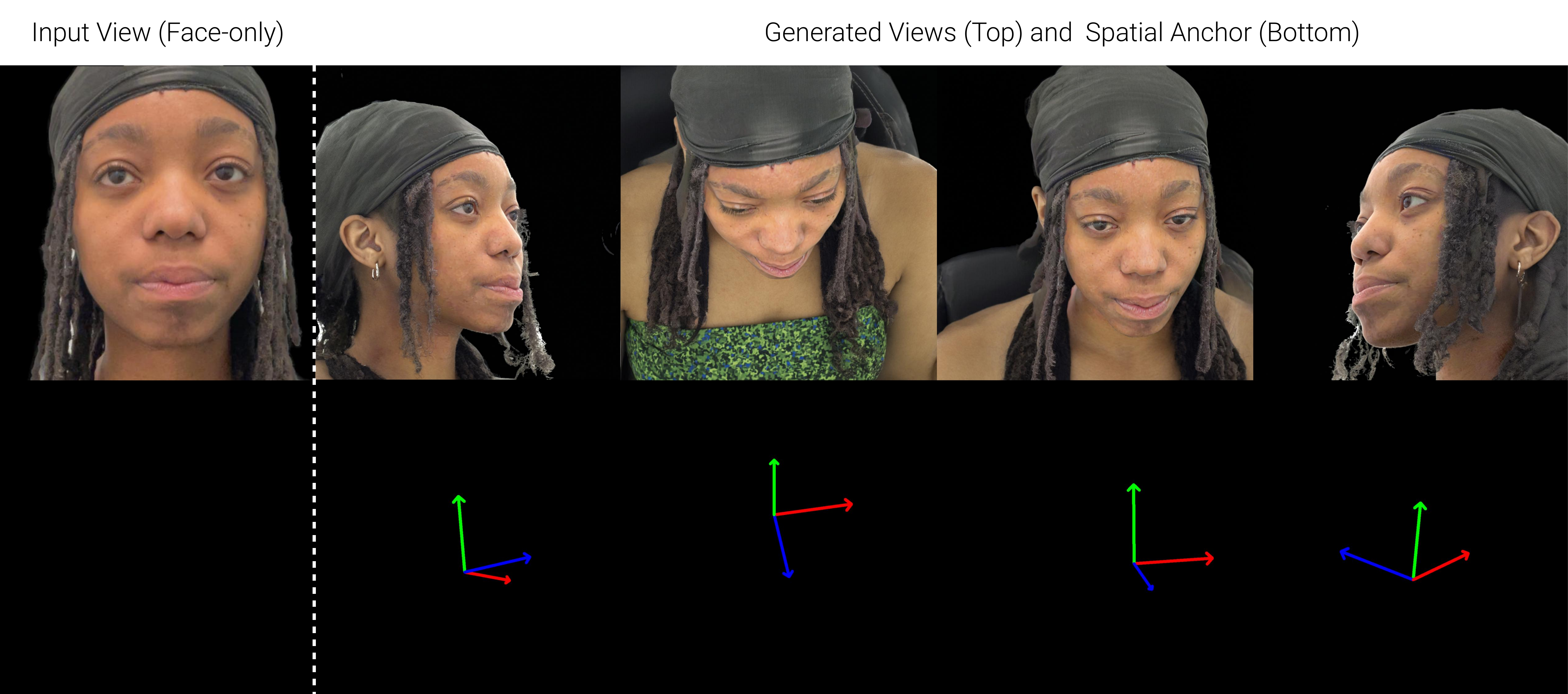}
    \caption{\textbf{Visualizing \spatialanchor}. Spatial anchor is an oriented 3D point in space, which helps anchor the generation by specifying a fixed headpose for generations. We put detailed discussion in ~\cref{sec:method}, ablation using it in ~\cref{tab:ablation} and more visuals on webpage.} 
    \label{fig:qual_extreme}
\end{figure*}

\end{document}